%% file: iclr2027_conference.tex
\documentclass{article} % For LaTeX2e
\usepackage{iclr2027_conference,times}

\input{math_commands.tex}

\input{commands.tex}
\usepackage{hyperref}
\usepackage{longtable}
\usepackage{url}
\usepackage{caption}
\usepackage{algorithmic}
\usepackage{tabularx}
\usepackage{booktabs}
\usepackage{amssymb}
\usepackage{xcolor}
\usepackage{tikz}
\usepackage{pgfplots}
\usepackage{wrapfig}
\usepackage{booktabs}
\usepackage{longtable}
\usepackage{hyperref}
\usepackage{url}
\usepackage{siunitx} % Recommended for decimal alignment
\usepackage{enumitem}
\usepgfplotslibrary{groupplots}
\usepackage[table]{xcolor}
\pgfplotsset{compat=1.18} % You can change this to 1.17 or 1.16 if your compiler complains
\usepackage{color}

\usepackage[ruled,vlined]{algorithm2e}
\SetKwInOut{Input}{Input}
\SetKwFor{For}{for}{do}{end}
\SetKwFor{While}{while}{do}{end}
\SetKwIF{If}{ElseIf}{Else}{if}{then}{else if}{else}{end}
\SetKw{KwRet}{return}

\title{You're Hired: Strategic Model Selection for LLM Collaboration}

\author{Zongwan Cao\thanks{\quad equal contribution} \ $^1$ \ \  Ziyuan Yang\footnotemark[1] \ $^1$ \ \  Shangbin Feng\footnotemark[1] \ $^1$ \ \ \\ \textbf{Michal Duan}$^2$ \ \ \textbf{Skyler Hallinan}$^2$ \ \ \textbf{Bingbing Wen}$^1$ \ \  \textbf{Lucy Lu Wang}$^{13}$ \ \ \textbf{Yulia Tsvetkov}$^1$ \\ $^1$University of Washington \ \ $^2$University of Southern California \ \ $^3$Allen Institute for AI \\ \texttt{\{zongwanc,ziyuan86\}@uw.edu} \ \ \texttt{shangbin@cs.washington.edu}}

\iclrfinalcopy % Uncomment for camera-ready version, but NOT for submission.
\begin{document}

\maketitle

\begin{abstract}
While multi-agent and model collaboration algorithms gain traction to combine the strengths of diverse Large Language Models (LLMs), existing systems remain bottlenecked on pre-defined and hand-crafted model pools. In this work, we investigate the problem of \emph{model selection in multi-LLM systems}. We propose and systematically evaluate a taxonomy of 9 selection algorithms ranging from diversity of model descriptions, capability-aware behavioral diversity, and LLM-based recruiters. We conduct extensive experiments across two candidate pools of 10 and 32 models, deployed in four model collaboration algorithms, and evaluated across tasks spanning math, coding, QA, and reasoning. Results demonstrate that successful selection algorithms greatly outperform random or heuristics-based teams such as merely selecting the models with top individual performance, by up to 36.1\% across settings. Specifically, capability- and training-based selection strategies alleviate selection variance and achieve the best performance, which we recommend to employ before deploying real-world multi-LLM systems.
Further analysis reveals that larger candidate pools pose greater challenges to shallow selection heuristics, while algorithms grounded in interacting with candidate models and understanding model capability robustly filter out misaligned, unsafe models, as well as generalizing to novel, out-of-distribution tasks. Together, we establish that principled and informed team selection is critical and present strong model selection algorithms for assembling effective multi-LLM systems.
\end{abstract}

\section{Introduction}

Recent advances in Large Language Models (LLMs) have fueled a transition from single-model inference to multi-model collaborative systems. Frameworks such as multi-agent debate \citep{du2024multiagentdebate}, dynamic routing \citep{feng2026moco,ong2025routellm}, and parameter fusion \citep{yu2023languagemodels,yadav2023ties} demonstrate that combining multiple LLMs can overcome individual biases \citep{liang2023divergent, zhao2025language}, reduce hallucinations \citep{du2024multiagentdebate}, and push the frontier of task performance\citep{feng2026moco,feng2026participation}. Yet, the effectiveness of these collaborative architectures rests on a critical, often overlooked assumption: \textbf{a well-composed team}. With over 2 million open language models available \citep{wolf2020transformers}, assembling a synergistic pool of collaborator models beyond manual cherry-picking remains an open challenge.

Currently, multi-LLM teams are typically assembled via ad-hoc heuristics, uniform random sampling, or by simply grouping the largest available models. Yet these practices overlook a central challenge: models that are individually strong or superficially diverse may not necessarily form an effective team \citep{kuncheva2003diversity, sanchez2025llm}. This motivates a fundamental question: \emph{How should we select models to form an effective collaborative team?}
In this work, we systematically investigate the science of model selection in multi-LLM systems. We formalize the candidate selection problem and investigate a comprehensive taxonomy of selection strategies that differ in the signals used to characterize candidate models and the procedures used to form teams. Specifically, we investigate and propose five categories of algorithms:
\emph{standard baselines} such as employing models with best solo scores, \emph{stated diversity} from model descriptions, \emph{capability-aware diversity} capturing observed behavioral patterns, \emph{hybrid} approaches ensembling description and capability diversity, as well as \emph{LLM-based recruiters}.

We evaluate these selectors across two candidate pools of varying heterogeneity with 10 and 32 models, diverse benchmarks, and four distinct collaboration mechanisms. Specifically, our evaluation encompasses tasks spanning mathematical problem solving, code generation, question answering, and complex reasoning. Our results reveal four key insights: First, randomly assembled teams underperform and introduce substantial performance variance, highlighting the necessity of fine-grained team selection; Second, capability-aware strategies frequently build stronger teams that reliably outperform the best single model and heuristics-based baselines; Third, capability-aware selection is the safest default as they consistently emerged as the top performers. If adversarial robustness is the priority, quality-filtered approaches (such as Agentic Top-$k$, Nested Diversity) are recommended to select out the malicious models.
Furthermore, we demonstrate that effective selection strategies scales robustly in redundant ecosystems and generalizes to out-of-distribution tasks. Ultimately, who you select for a multi-LLM team matters just as much as how they collaborate, and our algorithms provide a practical blueprint for hiring the right models for the job.

% Main figure

\begin{figure}[t]
\vspace{-1cm}
\centering
\includegraphics[width=1.0\linewidth]{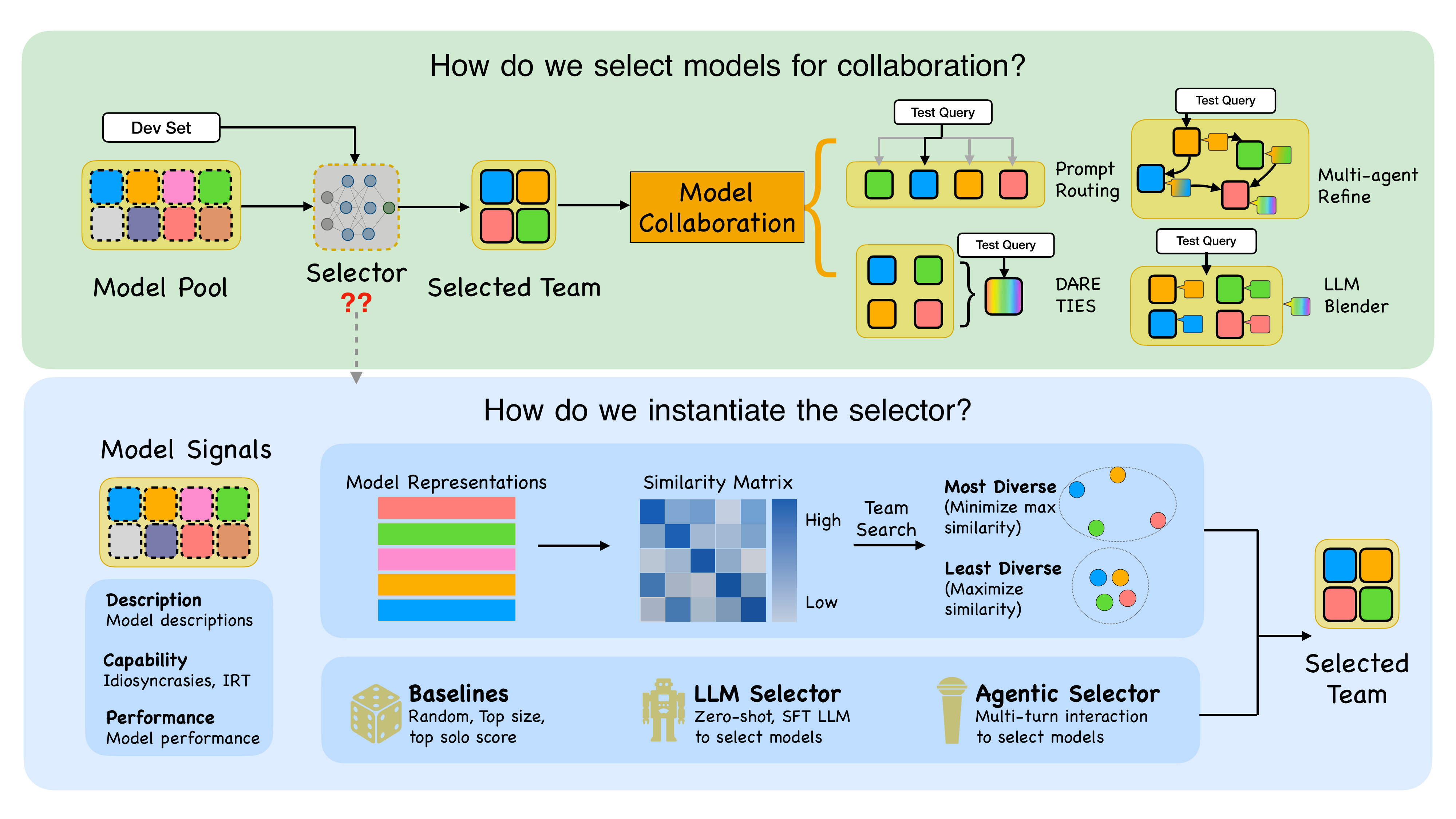}
\vspace{-20pt}
\caption{Overview of our work. Given a candidate model pool, we study how to select a team for downstream collaboration. Our selectors consist of standard baselines, stated diversity, capability-aware-diversity, hybrid approaches, and LLM-based recruiters. The selected models are then composed into a team and evaluated across diverse collaboration paradigms.}
\label{fig:main_overview}
\vspace{-0.2cm}
\end{figure}

\section{Related Work}
\paragraph{Multi-LLM Collaboration.} The paradigm of combining multiple language models takes several forms. Routing mechanisms aim to dynamically assign inputs to the most suitable expert \citep{ong2025routellm}. Generative fusion \citep{jiang2023llmblender} and multi-agent debate \citep{du2024multiagentdebate} allow models to iteratively critique and refine each other's outputs. At the parameter level, techniques like DARE \citep{yu2023languagemodels} and TIES \citep{yadav2023ties} merge the weights of distinct models into a single unified model. While these methods provide the infrastructure for collaboration, they generally assume the input ensemble is pre-determined, leaving the question of optimal team composition largely unaddressed.

\paragraph{Model Diversity and Selection.} Diversity is a key consideration in ensemble construction, as complementary models can reduce errors  \citep{kuncheva2003diversity}. Recent works shows that diverse expertise across LLM agents can improve collective performance \citep{zhang2025diversity},  while structural metadata can be used to characterize model relationships in the broader LLM ecosystem \citep{li2024transfergraph,laufer2025anatomy}. However, these structural proxies do not directly characterize behavioral complementarity. Our work bridges this gap by adapting combinatorial search algorithms to fine-grained, behavioral signals, ensuring that selected models are diverse and effective.
% in their actual capabilities rather than just their stated descriptions.

\paragraph{Capability Modeling.}
Quantifying the intrinsic abilities of LLMs has evolved beyond simple benchmark leaderboards \citep{liang2022helm}. Methods such as Item Response Theory (IRT), originally developed for scaling NLP evaluation \citep{lalor2016irt}, are recently adapted to map LLMs into continuous latent ability spaces based on item-level success patterns \citep{chen2025irt}. Similarly, attribution confusability \citep{sun2025idiosyncrasies} measures behavioral overlap by analyzing whether a classifier can distinguish between the text generated by different models. We leverage these advanced capability representations not for leaderboard ranking, but as the underlying signals for our diversity-seeking selection algorithms.

\section{Methodology}
\label{sec:methodology}

% \subsection{Problem Formulation}
% \label{sec:methodology:formulation}

\paragraph{Problem Formulation} Let $\mathcal{M} = \{m_1, \ldots, m_P\}$ denote a pool of $P$ candidate language models, 
% each equipped with a natural-language description $\textit{desc}(m)$ of its specialization. 
Given a target team size $n$, a selector $\sigma$ chooses a subset of models from the model pool to form team $T$:
\begin{equation}
    \sigma (\mathcal{M}, n) = T, \qquad T \subseteq \mathcal{M}, \; |T| = n,
\end{equation}

Given a collaboration method $c \in \mathcal{C}$ and a task $d \in \mathcal{D}$, the performance of the selected team $T$ under method $c$ on task $d$ is denoted by
% \begin{equation}
    % $\textit{score}: 2^{\mathcal{M}} \times \mathcal{C} \times \mathcal{D} \to [0,1],$
% \end{equation}
$\textit{score}(T, c, d)$.
% denotes the task performance of executing team $T$ under method $c$ on task $d$. 
In this work, we study how the choice of selector $\sigma$ affects $\textit{score}$ under different collaboration methods and tasks.

\begin{wraptable}{r}{0.5\textwidth}
%\vspace{-15pt}
\centering
\caption{Selection strategy taxonomy.}
\vspace{-5pt}
% Alg.~1 and Alg.~2 refer to exact diversity search and capability-seeded greedy search.}
\label{tab:selector-taxonomy}
\scriptsize

% --- Reduce space between columns (default is 6pt) ---
\setlength{\tabcolsep}{1pt} 

% The sum of \hsize values must equal the number of X columns (1.2 + 1.0 + 0.8 = 3)
\begin{tabularx}{\linewidth}{@{} 
    >{\raggedright\arraybackslash\hsize=0.85\hsize}X 
    >{\raggedright\arraybackslash\hsize=0.75\hsize}X 
    >{\raggedright\arraybackslash\hsize=0.7\hsize}X 
@{}}
\toprule[1.75pt]
\textbf{Selector $\sigma$} & \textbf{Signal $\mathcal{Z}$} & \textbf{Procedure} \\
\midrule
\multicolumn{3}{@{}l}{\textit{I. Standard Baselines}} \\
\hspace{1em} Random & None & Random sampling \\
\hspace{1em} Top size & Parameter count & Direct ranking \\
\hspace{1em} Top solo score & Solo performance & Direct ranking \\
% \addlinespace
\midrule
\multicolumn{3}{@{}l}{\textit{II. Stated Diversity}} \\
\hspace{1em} Description diversity & Model description & Exact (Alg.~\ref{alg:exact}) \\
% \addlinespace
\midrule
\multicolumn{3}{@{}l}{\textit{III. Capability-Aware Diversity}} \\
\hspace{1em} Idiosyncrasies & Attribution behavior & Greedy (Alg.~\ref{alg:greedy}) \\
\hspace{1em} IRT ability & IRT ability & Greedy (Alg.~\ref{alg:greedy}) \\
\hspace{1em} Performance profile & Benchmark profile & Exact (Alg.~\ref{alg:exact}) \\
% \addlinespace
\midrule
\multicolumn{3}{@{}l}{\textit{IV. Hybrid}} \\
\hspace{1em} Combined & Description + IRT  & Exact (Alg.~\ref{alg:exact}) \\
\hspace{1em} Nested & Description + IRT & Filter $+$ Alg.~\ref{alg:exact} \\
% \addlinespace
\midrule
\multicolumn{3}{@{}l}{\textit{V. LLM-based recruiters}} \\
\hspace{1em} LLM prompt & Model descriptions & Direct selection \\
\hspace{1em} SFT Classifier & Predicted team score & Team ranking \\
\hspace{1em} Agentic top-$k$ & Interview score & Direct ranking \\
\bottomrule[1.75pt]
\end{tabularx}
% \vspace{-8pt}
\end{wraptable}

\paragraph{Selection Strategies} Selecting a team $T$ requires information about candidate models that can inform the selection decision. We consider three common forms of such information: (i) \emph{stated} metadata, such as model cards \citep{mitchell2019modelcards}; (ii) \emph{demonstrated behavior} via item-level evaluation \citep{lalor2016irt,chen2025irt} or behavioral fingerprinting \citep{sun2025idiosyncrasies}; and (iii) external \emph{judgment} from strong LLMs or human raters \citep{zheng2023llmjudge}. Based on these observations, we instantiate our selector by two components: \textbf{Signal $\mathcal{Z}$} and \textbf{Procedure}. The signal captures candidate-level attributes, pairwise similarities, or whole-team judgments, while the procedure maps $\mathcal{Z}$ to team $T$ via ranking, search (Algorithms~\ref{alg:exact} and~\ref{alg:greedy}), filtering, or direct selection. Based on how these signals are constructed and used, we organize the resulting strategies into five families. Table~\ref{tab:selector-taxonomy} summarizes their signals and selection procedures.

% \begin{enumerate}[leftmargin=*, noitemsep, topsep=0pt]
%     \item \textbf{Signal ($\mathcal{Z}$):} Candidate-level attributes, pairwise similarities, or whole-team judgments.
%     \item \textbf{Procedure:} Maps $\mathcal{Z}$ to team $T$ via ranking, search (Algorithms~\ref{alg:exact} and \ref{alg:greedy}), filtering, or generation.
% \end{enumerate}

\subsection{Signal Construction}
\label{sec:methodology:signals}

Different selectors rely on different information about the candidate models. The information $\mathcal{Z}$ may include model descriptions, observed capabilities, or behavioral patterns. Each strategy constructs $\mathcal{Z}$ differently and then applies a corresponding selection procedure to form the team $T$.

For strategies that explicitly seek model diversity, we convert $\mathcal{Z}$ into a pairwise similarity matrix $\mathbf{S}=[s_{i,j}] \in \mathbb{R}^{P\times P}$, where $s_{i,j}$ measures the similarity between candidates $m_i$ and $m_j$ under the corresponding signal.

\paragraph{Baselines.} We consider three reference methods that do not explicitly model complementarity. \emph{Random} uses no signal. \emph{Top size} uses model parameter counts as the selection signal. \emph{Top solo score} uses performance on the development set of target task of each model as signal.

\paragraph{Stated diversity.}
We next use models' stated descriptions to characterize their different intended specializations. For \emph{Description diversity}, we define the selection signal as $\mathcal{Z}=\{\operatorname{desc}(m_i)\}_{i=1}^{P}$. We then encode description using a sentence transformer $\phi$, and construct the pairwise similarity matrix as $s_{i,j}=\cos(\phi(\operatorname{desc}(m_i),\phi(\operatorname{desc}(m_j))$.

\paragraph{Capability-aware diversity.} 
We characterize model diversity through observed capabilities and behavioral patterns.  \emph{Idiosyncrasies} and \emph{IRT} also derive behavioral signals from model responses on the development set of evaluation tasks, and \emph{Performance Profile} characterizes models on a held-out suite of benchmark tasks spanning math, code, QA, and reasoning, rather from the evaluation tasks.

\begin{itemize}[topsep=0pt, itemsep=0pt, leftmargin=16pt]

    \item \emph{Idiosyncrasies}: Following \citet{sun2025idiosyncrasies}, we train a text classifier to predict which candidate model generated a given response. Let $c_{i,j}$ denote the probability that the classifier predicts model $j$ for a response generated by model $i$. We use the resulting confusion probabilities as the selection signal $\mathcal{Z}=\{c_{i,j}\}_{i,j=1}^{P}$, and construct the pairwise similarity as $s{i,j}=\frac{1}{2}(c_{i,j}+c_{j,i})$. 
    
    \item \emph{IRT Ability}: Following \citet{chen2025irt}, we train an IRT model to estimate an ability vector $\theta_i$ for each candidate from its item-level successes and failures. We use these ability vectors as the selection signal $\mathcal{Z}=\{\theta_i\}_{i=1}^{P}$, and construct the pairwise similarity as $s_{i,j}=\cos(\theta_i,\theta_j)$.
    
    \item \emph{Performance Profile}: We represent each candidate model by its performance vector $p_i \in \mathbb{R}^{10}$ across held-out benchmarks. We use these performance vectors as the selection signal $\mathcal{Z}=\{p_i\}_{i=1}^{P}$, and construct the pairwise similarity as $s_{i,j}=\cos(p_i,p_j)$.    
\end{itemize}

% \paragraph{Hybrid.} We combine descriptions and IRT capabilities in two ways. For the \emph{Combined} signal, the description embedding $v_i$ and IRT ability embedding $\theta_i$ are concatenated into a joint representation $h_i = [v_i; \theta_i]$. Pairwise similarity is evaluated in this joint space: $s_{i,j} = \cos(h_i, h_j)$.  \emph{Nested} relies on these same signals, applying them sequentially rather than combining them into a single representation.

\paragraph{Hybrid.}
We combine signals from \emph{Description Diversity} and \emph{IRT Ability} in two ways. For \emph{Combined}, we concatenate the description embedding $\operatorname{desc}(m_i)$ and IRT ability embedding $\theta_i$ into a joint representation $h_i=[v_i;\theta_i]$. We use these joint representations as the selection signal, $\mathcal{Z}=\{h_i\}_{i=1}^{P}$ and compute $s_{i,j}=\cos(h_i,h_j)$. \emph{Nested} retains the description and IRT signals separately and use them sequentially.

\paragraph{LLM-based recruiters.}
Finally, rather than constructing a fixed pairwise similarity matrix $\mathbf{S}$, these methods use learned or LLM-elicited judgments as selection signals.
For \emph{LLM prompt}, the signal consists of model descriptions $\mathcal{Z}=\{\operatorname{desc}(m_i)\}_{i=1}^{P}$.
For \emph{SFT Classifier}, $\mathcal{Z}$ consists of predicted team scores produced by a learned regression model from the descriptions of the team members.
For \emph{Agentic top-$k$}, the signal consists of scalar interview scores, $\mathcal{Z}=\{q_i\}_{i=1}^{P}$, where $q_i$ is assigned to candidate $m_i$ by an LLM recruiter.
We provide details in Appendix~\ref{app:learned-selectors}.

\vspace{-1mm}
\subsection{Selection Procedures}
\label{sec:methodology:procedures}

Given a selection signal $\mathcal{Z}$, each selector applies a corresponding procedure to form a team $T$ of size $n$. Depending on the selector, this procedure takes the form of direct sampling or ranking, diversity-based search, sequential filtering, or LLM-based selection.

\paragraph{Baselines.}
The baseline selectors require no complex search. \emph{Random} uniformly samples a team of $n$ candidates from model pool. \emph{Top size} and \emph{Top solo score} directly rank candidates by their respective scalar signals and select the top $n$ candidates.

\paragraph{Diversity-based selection.}
For selectors that construct a pairwise similarity matrix $\mathbf{S}$, we define team diversity using a max--min dispersion criterion, with average pairwise similarity as a tie-breaker. Specifically, we define the \emph{most} diverse team as the one that minimizes the maximum pairwise similarity, whereas the \emph{least} diverse team maximizes it. We additionally compare against average-dispersion objectives in Appendix~\ref{app:maxmin-vs-avg}. We consider two selection procedures:

\begin{itemize}[topsep=0pt, itemsep=0pt, leftmargin=16pt]
\item \emph{Exact diversity search} (Algorithm~\ref{alg:exact}): We enumerate all candidate teams of size $n$ and select a team according to the diversity criterion. We use this procedure for \emph{Description Diversity} and \emph{Performance Profile}, where pairwise similarity provides a direct diversity signal.

\item \emph{Capability-seeded greedy search} (Algorithm~\ref{alg:greedy}): We incorporate candidate capability into the search and use this procedure for \emph{Idiosyncrasies} and \emph{IRT Ability}, where weak models may produce unusual errors and therefore appear distinctive. The search starts from the two highest-performing candidates and greedily adds the remaining members according to the diversity criterion, reducing the risk of selecting weak models solely for their distinctiveness.
\end{itemize}

\paragraph{Hybrid.}
These methods combine the signals above in two ways. \emph{Combined} applies Algorithm~\ref{alg:exact} to the joint representation $h_i$. \emph{Nested} first retains the stronger half of the candidate pool based on \emph{IRT Ability} scores $\theta_i$, then applies Algorithm~\ref{alg:exact} to the remaining candidates using the description-based similarity signal.

\paragraph{LLM-based recruiters.}
These procedures bypass pairwise similarity matrices and select teams using learned or zero-shot LLM recruiters: \emph{LLM prompt}: directly selects a team from candidate descriptions using a zero-shot LLM recruiter. \emph{SFT Classifier}: enumerates all teams of size $n$, scores each with a fine-tuned LLM-based regressor, and selects the highest-scoring team. \emph{Agentic top-$k$}: conducts multi-turn LLM interviews with each candidate, ranks candidates by their interview scores, and selects the top $n$ models. We provide further details of these procedures in Appendix~\ref{app:learned-selectors}.

\begin{algorithm}[t!]
    \caption{Exact Diversity Search}
    \label{alg:exact}
    \scriptsize
    \begin{algorithmic}[1]
    \REQUIRE Similarity matrix $\mathbf{S} \in \mathbb{R}^{P\times P}$, team size $n$, mode $\in \{\textsc{most}, \textsc{least}\}$
    \STATE $T^\star \gets \varnothing$
    \FOR{each $T \subseteq \{1,\ldots,P\}$ with $\vert{}T\vert{} = n$}
        \STATE $s_{\text{max}}(T) \gets \max_{i,j \in T,\, i \neq j} s_{i,j}$
        \STATE $s_{\text{avg}}(T) \gets 1/\binom{n}{2} \sum_{i,j \in T,\, i < j} s_{i,j}$
        \STATE $s(T) \gets (s_{\text{max}}(T), s_{\text{avg}}(T))$
        \IF{mode $= \textsc{most}$ \textbf{and} $s(T) < s(T^\star)$}
            \STATE $T^\star \gets T$
        \ELSIF{mode $= \textsc{least}$ \textbf{and} $s(T) > s(T^\star)$}
            \STATE $T^\star \gets T$
        \ENDIF
    \ENDFOR
    \RETURN $T^\star$
    \end{algorithmic}
\end{algorithm}

\begin{algorithm}[t!]
    \caption{Capability-Seeded Greedy Search}
    \label{alg:greedy}
    \scriptsize
    \begin{algorithmic}[1]
    \REQUIRE Similarity matrix $\mathbf{S} \in \mathbb{R}^{P\times P}$, team size $n$, scores $\{\textit{cap}(i)\}$, mode $\in \{\textsc{most}, \textsc{least}\}$
    \STATE $\textit{seed} \gets$ top-2 indices by $\textit{cap}(\cdot)$
    \STATE $T \gets \textit{seed}$
    \STATE $R \gets \{1,\ldots,P\} \setminus T$
    \WHILE{$\vert{}T\vert{} < n$}
        \FOR{$c \in R$}
            \STATE $s(c) \gets \max_{j \in T} s_{c, j}$
        \ENDFOR
        \STATE $c^\star \gets \arg\min_{c \in R} s(c)$ \textbf{if} mode $=\textsc{most}$ \textbf{else} $\arg\max_{c \in R} s(c)$
        \STATE $T \gets T \cup \{c^\star\}$
        \STATE $R \gets R \setminus \{c^\star\}$
    \ENDWHILE
    \RETURN $T$
    \end{algorithmic}
\end{algorithm}

\begin{table*}[t]
\centering
% Define custom distinct RGB colors for the heatmap (ColorBrewer Greens)
\definecolor{lightG}{HTML}{EEFCDE}    % Tier 1: Very pale green
\definecolor{darkG}{HTML}{89C492}     % Tier 2: Medium clear green
\definecolor{darkestG}{HTML}{238B45}  % Tier 3: Vibrant bold green
\newcommand{\lightG}{\cellcolor{lightG}}
\newcommand{\darkG}{\cellcolor{darkG}}
\newcommand{\darkestG}{\cellcolor{darkestG}}

\vspace{-10pt}
\caption{Main collaborative performance results across all datasets and mechanisms for \textbf{Pool 1} (top) and \textbf{Pool 2} (bottom). The highest mean score in each column per pool is \textbf{bolded}. Cells are colored to highlight the efficacy of adaptive selection based on mean performance: \colorbox{lightG}{Light Green} indicates the team beats the \emph{Random} baseline, \colorbox{darkG}{Medium Green} indicates the team beats all standard baselines (\emph{Random}, \emph{Top Solo}, and \emph{Top Size} when applicable), and \colorbox{darkestG}{\textcolor{white}{Dark Green}} indicates the team beats all standard baselines plus the \emph{Single Best Model}.}
\label{tab:main_results}

% ==========================================
% POOL 1 TABULAR
% ==========================================
%\vspace{0.5em}
\textbf{Pool 1 (Homogeneous Candidates, 10 $\rightarrow$ 4 models)} \\
\vspace{0.2em}
\resizebox{\textwidth}{!}{%
\begin{tabular}{@{}l ccc ccc ccc ccc@{}}
\toprule[1.75pt]
& \multicolumn{3}{c}{\textbf{Prompt Routing}} & \multicolumn{3}{c}{\textbf{Multiagent Refine}} & \multicolumn{3}{c}{\textbf{DARE-TIES}} & \multicolumn{3}{c}{\textbf{LLM-Blender}} \\
\cmidrule(lr){2-4} \cmidrule(lr){5-7} \cmidrule(lr){8-10} \cmidrule(l){11-13}
\textbf{Selection Method} & GSM8K & TQA & MBPP & GSM8K & TQA & MBPP & GSM8K & TQA & MBPP & GSM8K & TQA & MBPP \\
\midrule[0.75pt]

\multicolumn{13}{@{}l}{\textbf{Standard Baselines}} \\
\quad \emph{Single Best Model} & \textbf{65.6}$_{\pm 1.5}$ & 59.3$_{\pm 2.0}$ & 47.0$_{\pm 2.3}$ & 65.6$_{\pm 1.5}$ & \textbf{59.3}$_{\pm 2.0}$ & 47.0$_{\pm 2.3}$ & 65.6$_{\pm 1.5}$ & 59.3$_{\pm 2.0}$ & 47.0$_{\pm 2.3}$ & 65.6$_{\pm 1.5}$ & 59.3$_{\pm 2.0}$ & 47.0$_{\pm 2.3}$ \\

% \midrule
% \multicolumn{13}{@{}l}{\textbf{Group 2: Standard Baselines}} \\
\quad \emph{Random} & $50.5_{\pm 10.0}$ & $48.0_{\pm 5.7}$ & $51.7_{\pm 6.5}$ & $45.4_{\pm 2.6}$ & $42.8_{\pm 6.1}$ & $42.7_{\pm 4.4}$ & $45.2_{\pm 13.4}$ & $59.8_{\pm 4.7}$ & $52.4_{\pm 7.5}$ & $59.8_{\pm 7.1}$ & $48.0_{\pm 5.4}$ & $53.8_{\pm 7.3}$ \\
\quad \emph{Top Solo Score} & $62.8_{\pm 1.5}$ & $43.9_{\pm 2.0}$ & $45.2_{\pm 2.2}$ & $60.2_{\pm 1.6}$ & $50.2_{\pm 2.0}$ & $36.1_{\pm 2.2}$ & $31.4_{\pm 1.5}$ & $63.9_{\pm 1.9}$ & $46.4_{\pm 2.2}$ & \textbf{73.9}$_{\pm 1.4}$ & $51.9_{\pm 2.0}$ & $47.0_{\pm 2.3}$ \\

\midrule[0.75pt]
% \multicolumn{13}{@{}l}{\textbf{Group 3: Advanced Selectors}} \\
\multicolumn{13}{@{}l}{\textbf{Stated Diversity}} \\
\quad \emph{Description (least)} & $26.4_{\pm 1.4}$ & $34.2_{\pm 1.9}$ & \darkestG $54.6_{\pm 2.3}$ & \lightG $51.0_{\pm 1.6}$ & \lightG $43.0_{\pm 2.0}$ & \darkestG $47.4_{\pm 2.2}$ & $34.4_{\pm 1.5}$ & \lightG $61.8_{\pm 1.9}$ & \darkestG $59.1_{\pm 2.3}$ & $49.7_{\pm 1.6}$ & \lightG $49.1_{\pm 2.0}$ & \darkestG $55.4_{\pm 2.3}$ \\
\quad \emph{Description (most)} & \darkG $63.4_{\pm 1.5}$ & \darkestG \textbf{60.0}$_{\pm 2.0}$ & \darkestG $57.7_{\pm 2.3}$ & \lightG $50.0_{\pm 1.6}$ & $40.7_{\pm 2.0}$ & $38.8_{\pm 2.2}$ & \darkG $54.0_{\pm 1.6}$ & \darkestG \textbf{66.1}$_{\pm 1.9}$ & \darkestG $56.9_{\pm 2.2}$ & $48.5_{\pm 1.6}$ & \darkestG \textbf{62.1}$_{\pm 2.0}$ & \darkestG $57.5_{\pm 2.2}$ \\
\midrule[0.75pt]
\multicolumn{13}{@{}l}{\textbf{Capability-Aware Diversity}} \\
\quad \emph{Idiosyncrasies (least)} & \lightG $56.5_{\pm 1.6}$ & $40.5_{\pm 2.0}$ & \darkestG $56.3_{\pm 2.3}$ & \lightG $53.3_{\pm 1.6}$ & $36.3_{\pm 1.9}$ & \darkestG \textbf{49.7}$_{\pm 2.2}$ & \darkG $58.4_{\pm 1.6}$ & $56.9_{\pm 2.0}$ & \darkestG $54.6_{\pm 2.2}$ & \lightG $67.6_{\pm 1.5}$ & \darkG $52.7_{\pm 2.0}$ & $49.3_{\pm 2.3}$ \\
\quad \emph{Idiosyncrasies (most)} & \lightG $51.1_{\pm 1.6}$ & $40.7_{\pm 2.0}$ & \darkestG $56.7_{\pm 2.2}$ & $41.0_{\pm 1.6}$ & \lightG $44.9_{\pm 2.0}$ & $38.6_{\pm 2.2}$ & \darkG $59.1_{\pm 1.6}$ & \lightG $63.9_{\pm 1.9}$ & \darkestG $55.4_{\pm 2.2}$ & \lightG $70.9_{\pm 1.4}$ & \lightG $51.5_{\pm 2.0}$ & $53.0_{\pm 2.3}$ \\
\quad \emph{IRT-Net (least)} & \lightG $60.6_{\pm 1.5}$ & \darkG $56.9_{\pm 2.0}$ & $33.3_{\pm 2.1}$ & \darkestG $72.2_{\pm 1.4}$ & \lightG $47.6_{\pm 2.0}$ & $25.5_{\pm 2.0}$ & \darkG $57.5_{\pm 1.6}$ & \lightG $60.1_{\pm 2.0}$ & $43.1_{\pm 2.2}$ & \lightG $61.8_{\pm 1.5}$ & \darkG $55.6_{\pm 2.0}$ & $47.4_{\pm 2.2}$ \\
\quad \emph{IRT-Net (most)} & $38.2_{\pm 1.5}$ & \darkG $53.6_{\pm 2.0}$ & \darkestG $57.5_{\pm 2.2}$ & \darkG $60.8_{\pm 1.6}$ & \lightG $46.0_{\pm 2.0}$ & \darkG $45.8_{\pm 2.3}$ & \darkG $50.2_{\pm 1.6}$ & \lightG $63.0_{\pm 1.9}$ & \darkestG $58.9_{\pm 2.2}$ & \lightG $69.1_{\pm 1.5}$ & \darkG $56.2_{\pm 2.0}$ & $36.3_{\pm 2.1}$ \\
\quad \emph{Perf.\ Profile (least)} & \darkG $64.9_{\pm 1.5}$ & \darkestG \textbf{60.0}$_{\pm 2.0}$ & \darkestG $58.1_{\pm 2.3}$ & \lightG $56.5_{\pm 1.6}$ & $41.8_{\pm 2.0}$ & \darkG $45.0_{\pm 2.2}$ & \darkG $45.3_{\pm 1.6}$ & \darkestG $64.5_{\pm 1.9}$ & \darkestG \textbf{60.4}$_{\pm 2.2}$ & $50.1_{\pm 1.6}$ & \darkG $58.7_{\pm 2.0}$ & \darkestG $57.1_{\pm 2.2}$ \\
\quad \emph{Perf.\ Profile (most)} & $37.2_{\pm 1.5}$ & $46.5_{\pm 2.0}$ & \darkestG $56.5_{\pm 2.3}$ & $12.2_{\pm 1.1}$ & $30.0_{\pm 1.9}$ & $42.5_{\pm 2.2}$ & \darkG $58.2_{\pm 1.6}$ & $59.2_{\pm 2.0}$ & \darkestG $54.0_{\pm 2.3}$ & \lightG $60.5_{\pm 1.6}$ & \lightG $50.6_{\pm 2.0}$ & $53.4_{\pm 2.3}$ \\
\midrule[0.75pt]
\multicolumn{13}{@{}l}{\textbf{Hybrid}} \\
\quad \emph{Combined (least)} & \lightG $62.1_{\pm 1.5}$ & \darkG $52.0_{\pm 2.0}$ & $33.9_{\pm 2.1}$ & \darkestG $70.4_{\pm 1.4}$ & \lightG $44.4_{\pm 2.0}$ & $24.9_{\pm 2.0}$ & \darkG $59.7_{\pm 1.5}$ & \lightG $62.2_{\pm 1.9}$ & $47.0_{\pm 2.3}$ & \lightG $70.5_{\pm 1.5}$ & \darkG $58.4_{\pm 2.0}$ & $33.3_{\pm 2.1}$ \\
\quad \emph{Combined (most)} & \lightG $56.4_{\pm 1.6}$ & \darkG $49.3_{\pm 2.0}$ & \darkestG $56.1_{\pm 2.2}$ & \lightG $49.9_{\pm 1.6}$ & $38.7_{\pm 2.0}$ & \darkG $45.6_{\pm 2.3}$ & \darkestG \textbf{67.5}$_{\pm 1.5}$ & \lightG $62.7_{\pm 2.0}$ & \darkestG $59.3_{\pm 2.2}$ & $59.0_{\pm 1.5}$ & \darkG $56.9_{\pm 2.0}$ & \darkestG \textbf{58.3}$_{\pm 2.2}$ \\
\quad \emph{Nested (least)} & \darkG $63.2_{\pm 1.6}$ & \darkG $48.3_{\pm 2.0}$ & $37.8_{\pm 2.2}$ & $40.9_{\pm 1.6}$ & $41.0_{\pm 2.0}$ & $34.5_{\pm 2.1}$ & \darkG $51.6_{\pm 1.6}$ & $58.2_{\pm 2.0}$ & $47.2_{\pm 2.3}$ & \lightG $72.7_{\pm 1.4}$ & \darkG $55.1_{\pm 2.0}$ & $31.6_{\pm 2.1}$ \\
\quad \emph{Nested (most)} & \lightG $61.4_{\pm 1.5}$ & \darkG $53.8_{\pm 2.0}$ & $28.7_{\pm 2.1}$ & \darkestG \textbf{72.9}$_{\pm 1.4}$ & \lightG $43.1_{\pm 2.0}$ & $26.1_{\pm 2.0}$ & \darkG $56.6_{\pm 1.6}$ & $59.0_{\pm 2.0}$ & $47.6_{\pm 2.3}$ & \lightG $71.2_{\pm 1.4}$ & \darkG $55.1_{\pm 2.0}$ & $31.6_{\pm 2.1}$ \\
\midrule[0.75pt]
\multicolumn{13}{@{}l}{\textbf{LLM-based recruiter}} \\
\quad \emph{LLM Prompt} & $40.1_{\pm 1.5}$ & $43.9_{\pm 2.0}$ & $39.4_{\pm 2.2}$ & \lightG $58.7_{\pm 1.6}$ & \lightG $47.0_{\pm 2.0}$ & $41.5_{\pm 2.2}$ & \darkG $54.5_{\pm 1.6}$ & $53.2_{\pm 2.0}$ & \darkestG $54.2_{\pm 2.3}$ & \lightG $70.0_{\pm 1.4}$ & \lightG $48.9_{\pm 2.0}$ & $36.5_{\pm 2.2}$ \\
\quad \emph{SFT Classifier} & $44.7_{\pm 1.6}$ & \darkG $51.0_{\pm 2.0}$ & \darkestG \textbf{60.0}$_{\pm 2.2}$ & $20.7_{\pm 1.3}$ & $35.3_{\pm 1.9}$ & \darkG $46.2_{\pm 2.3}$ & \darkG $63.5_{\pm 1.5}$ & \lightG $61.6_{\pm 1.9}$ & \darkestG $54.8_{\pm 2.2}$ & $39.1_{\pm 1.5}$ & \darkG $54.0_{\pm 2.0}$ & \darkestG $56.9_{\pm 2.2}$ \\
\quad \emph{Agentic Top-K} & \darkG $63.4_{\pm 1.5}$ & \darkG $59.0_{\pm 2.0}$ & $39.4_{\pm 2.2}$ & \darkG $63.2_{\pm 1.5}$ & \lightG $43.9_{\pm 2.0}$ & $36.8_{\pm 2.2}$ & \darkG $63.7_{\pm 1.5}$ & \lightG $61.4_{\pm 2.0}$ & \darkestG $54.4_{\pm 2.3}$ & \lightG $72.5_{\pm 1.4}$ & \darkG $57.0_{\pm 2.0}$ & $37.4_{\pm 2.2}$ \\
\bottomrule[1.75pt]
\end{tabular}
}

% ==========================================
% POOL 2 TABULAR
% ==========================================
\vspace{0.5em} % Adds space between the two tables
\textbf{Pool 2 (Heterogeneous Candidates, 32 $\rightarrow$ 4 models)} \\
\vspace{0.2em}
\resizebox{\textwidth}{!}{%
\begin{tabular}{@{}l ccc ccc ccc ccc@{}}
\toprule[1.75pt]
& \multicolumn{3}{c}{\textbf{Prompt Routing}} & \multicolumn{3}{c}{\textbf{Multiagent Refine}} & \multicolumn{3}{c}{\textbf{DARE-TIES}} & \multicolumn{3}{c}{\textbf{LLM-Blender}} \\
\cmidrule(lr){2-4} \cmidrule(lr){5-7} \cmidrule(lr){8-10} \cmidrule(l){11-13}
\textbf{Selection Method} & GSM8K & TQA & MBPP & GSM8K & TQA & MBPP & GSM8K & TQA & MBPP & GSM8K & TQA & MBPP \\
\midrule[0.75pt]

\multicolumn{13}{@{}l}{\textbf{Standard Baselines}} \\
\quad \emph{Single Best Model} & \textbf{82.9}$_{\pm 1.0}$ & $46.0_{\pm 2.0}$ & $55.9_{\pm 2.2}$ & $82.9_{\pm 1.0}$ & $46.0_{\pm 2.0}$ & $55.9_{\pm 2.2}$ & $82.9_{\pm 1.0}$ & $46.0_{\pm 2.0}$ & $55.9_{\pm 2.2}$ & $82.9_{\pm 1.0}$ & $46.0_{\pm 2.0}$ & $55.9_{\pm 2.2}$ \\

% \midrule
% \multicolumn{13}{@{}l}{\textbf{Group 2: Standard Baselines}} \\
\quad \emph{Random} & $49.6_{\pm 24.2}$ & $58.0_{\pm 5.7}$ & $53.1_{\pm 2.7}$ & $69.8_{\pm 16.4}$ & $45.4_{\pm 7.2}$ & $50.8_{\pm 5.3}$ & $28.4_{\pm 21.8}$ & $60.3_{\pm 3.4}$ & $61.4_{\pm 1.2}$ & $37.5_{\pm 25.3}$ & $43.0_{\pm 20.6}$ & $48.6_{\pm 17.5}$ \\
\quad \emph{Top Solo Score} & $78.9_{\pm 1.3}$ & $46.5_{\pm 2.0}$ & $58.1_{\pm 2.2}$ & $83.6_{\pm 1.2}$ & $24.8_{\pm 1.7}$ & $55.4_{\pm 2.3}$ & $85.1_{\pm 1.1}$ & $41.8_{\pm 2.0}$ & $60.2_{\pm 2.2}$ & $85.5_{\pm 1.1}$ & $42.3_{\pm 2.0}$ & $55.9_{\pm 2.3}$ \\
\quad \emph{Top Size} & $23.9_{\pm 1.4}$ & $58.7_{\pm 2.0}$ & $52.8_{\pm 2.3}$ & $74.6_{\pm 1.4}$ & $57.9_{\pm 2.0}$ & $48.5_{\pm 2.3}$ & $14.4_{\pm 1.1}$ & $65.6_{\pm 1.9}$ & $60.8_{\pm 2.2}$ & $35.5_{\pm 1.5}$ & $59.6_{\pm 2.0}$ & $60.0_{\pm 2.2}$ \\

\midrule[0.75pt]
% \multicolumn{13}{@{}l}{\textbf{Group 3: Advanced Selectors}} \\
\multicolumn{13}{@{}l}{\textbf{Stated Diversity}} \\
\quad \emph{Description (least)} & $8.6_{\pm 0.9}$ & $45.5_{\pm 2.0}$ & $47.2_{\pm 2.3}$ & $29.1_{\pm 1.4}$ & \lightG $51.7_{\pm 2.0}$ & \lightG $55.0_{\pm 2.3}$ & $11.4_{\pm 1.0}$ & $57.0_{\pm 2.0}$ & $60.8_{\pm 2.2}$ & $35.3_{\pm 1.5}$ & \lightG $55.9_{\pm 2.0}$ & \lightG $58.3_{\pm 2.3}$ \\
\quad \emph{Description (most)} & $21.7_{\pm 1.3}$ & \darkestG \textbf{63.9}$_{\pm 1.9}$ & \darkestG \textbf{62.4}$_{\pm 2.2}$ & \darkestG \textbf{87.9}$_{\pm 1.0}$ & \darkestG $59.5_{\pm 2.0}$ & \lightG $53.4_{\pm 2.3}$ & $16.1_{\pm 1.2}$ & \darkestG \textbf{66.8}$_{\pm 1.9}$ & \darkestG $63.0_{\pm 2.2}$ & $21.3_{\pm 1.3}$ & \darkestG \textbf{70.3}$_{\pm 1.9}$ & \darkestG $61.8_{\pm 2.2}$ \\
\midrule[0.75pt]
\multicolumn{13}{@{}l}{\textbf{Capability-Aware Diversity}} \\
\quad \emph{Idiosyncrasies (least)} & \lightG $77.6_{\pm 1.3}$ & $43.6_{\pm 2.0}$ & $58.1_{\pm 2.2}$ & \darkestG $85.4_{\pm 1.1}$ & $32.1_{\pm 1.9}$ & \darkestG $\bold{60.8}_{\pm 2.2}$ & \darkestG $85.6_{\pm 1.1}$ & $39.4_{\pm 2.0}$ & \darkestG \textbf{64.3}$_{\pm 2.2}$ & \darkestG \textbf{86.8}$_{\pm 1.1}$ & $40.5_{\pm 2.0}$ & $46.2_{\pm 2.3}$ \\
\quad \emph{Idiosyncrasies (most)} & \darkG $79.4_{\pm 1.3}$ & $46.7_{\pm 2.0}$ & \darkestG $59.1_{\pm 2.2}$ & \darkestG $83.9_{\pm 1.2}$ & $39.9_{\pm 2.0}$ & \lightG $51.9_{\pm 2.3}$ & \lightG $37.2_{\pm 1.5}$ & $57.2_{\pm 2.0}$ & $60.2_{\pm 2.2}$ & \darkestG $86.4_{\pm 1.1}$ & \lightG $50.9_{\pm 2.0}$ & \lightG $60.0_{\pm 2.2}$ \\
\quad \emph{IRT-Net (least)} & \lightG $66.0_{\pm 1.5}$ & $45.2_{\pm 2.0}$ & $43.3_{\pm 2.3}$ & \lightG $82.7_{\pm 1.2}$ & $19.0_{\pm 1.6}$ & $35.3_{\pm 2.2}$ & \darkestG \textbf{87.3}$_{\pm 1.1}$ & $34.8_{\pm 1.9}$ & \darkestG $62.6_{\pm 2.2}$ & \lightG $84.5_{\pm 1.1}$ & $35.7_{\pm 1.9}$ & $47.4_{\pm 2.3}$ \\
\quad \emph{IRT-Net (most)} & \lightG $77.3_{\pm 1.3}$ & $45.2_{\pm 2.0}$ & \lightG $56.9_{\pm 2.3}$ & \darkestG $84.3_{\pm 1.1}$ & $57.9_{\pm 2.0}$ & \lightG $54.4_{\pm 2.3}$ & \lightG $43.2_{\pm 1.6}$ & $59.6_{\pm 2.0}$ & \darkestG $62.0_{\pm 2.2}$ & \darkestG $86.6_{\pm 1.1}$ & \lightG $57.2_{\pm 2.0}$ & \lightG $57.3_{\pm 2.3}$ \\
\quad \emph{Perf.\ Profile (least)} & $14.7_{\pm 1.1}$ & \darkestG $60.0_{\pm 2.0}$ & \darkestG $59.8_{\pm 2.2}$ & $58.7_{\pm 1.6}$ & \lightG $48.1_{\pm 2.0}$ & $39.2_{\pm 2.2}$ & $12.7_{\pm 1.0}$ & \lightG $65.0_{\pm 1.9}$ & \darkestG $62.8_{\pm 2.2}$ & $26.9_{\pm 1.4}$ & \lightG $51.7_{\pm 2.0}$ & \lightG $60.0_{\pm 2.2}$ \\
\quad \emph{Perf.\ Profile (most)} & \lightG $77.5_{\pm 1.3}$ & $46.0_{\pm 2.0}$ & \lightG $54.2_{\pm 2.3}$ & \lightG $80.6_{\pm 1.2}$ & $40.5_{\pm 1.9}$ & \lightG $51.5_{\pm 2.3}$ & \lightG $53.9_{\pm 1.6}$ & $57.7_{\pm 2.0}$ & \darkestG $62.2_{\pm 2.2}$ & \lightG $81.8_{\pm 1.2}$ & \lightG $45.2_{\pm 2.0}$ & $30.4_{\pm 2.1}$ \\
\midrule[0.75pt]
\multicolumn{13}{@{}l}{\textbf{Hybrid}} \\
\quad \emph{Combined (least)} & $14.1_{\pm 1.1}$ & $45.4_{\pm 2.0}$ & $51.1_{\pm 2.3}$ & \lightG $78.0_{\pm 1.3}$ & \lightG $47.8_{\pm 2.0}$ & $50.1_{\pm 2.3}$ & $10.4_{\pm 1.0}$ & $58.4_{\pm 2.0}$ & $60.4_{\pm 2.2}$ & $19.3_{\pm 1.3}$ & \lightG $53.0_{\pm 2.0}$ & \darkestG $61.2_{\pm 2.2}$ \\
\quad \emph{Combined (most)} & \lightG $65.7_{\pm 1.5}$ & $46.5_{\pm 2.0}$ & $49.9_{\pm 2.3}$ & \lightG $81.5_{\pm 1.2}$ & $36.5_{\pm 1.9}$ & \lightG $53.6_{\pm 2.3}$ & \lightG $28.7_{\pm 1.4}$ & $57.9_{\pm 2.0}$ & \darkestG $61.6_{\pm 2.2}$ & \lightG $61.1_{\pm 1.5}$ & $33.6_{\pm 1.9}$ & \lightG $59.8_{\pm 2.2}$ \\
\quad \emph{Nested (least)} & \lightG $67.0_{\pm 1.5}$ & $44.7_{\pm 2.0}$ & $48.0_{\pm 2.3}$ & \darkestG $86.9_{\pm 1.1}$ & $34.8_{\pm 1.9}$ & \darkestG $56.7_{\pm 2.2}$ & \lightG $64.1_{\pm 1.5}$ & $59.5_{\pm 2.0}$ & $60.8_{\pm 2.2}$ & \darkestG $86.6_{\pm 1.1}$ & \lightG $50.1_{\pm 2.0}$ & \lightG $58.3_{\pm 2.2}$ \\
\quad \emph{Nested (most)} & $20.9_{\pm 1.3}$ & \darkestG $62.9_{\pm 2.0}$ & \darkestG $58.7_{\pm 2.2}$ & \lightG $82.5_{\pm 1.2}$ & \darkestG \textbf{65.0}$_{\pm 1.9}$ & \darkestG $58.7_{\pm 2.2}$ & $18.2_{\pm 1.2}$ & \lightG $65.1_{\pm 1.9}$ & \darkestG $63.9_{\pm 2.1}$ & \lightG $46.7_{\pm 1.6}$ & \darkestG $64.3_{\pm 1.9}$ & \darkestG \textbf{64.3}$_{\pm 2.1}$ \\
\midrule[0.75pt]
\multicolumn{13}{@{}l}{\textbf{LLM-based recruiters}} \\
\quad \emph{LLM Prompt} & \lightG $76.2_{\pm 1.3}$ & $50.6_{\pm 2.0}$ & \lightG $57.3_{\pm 2.2}$ & \lightG $80.6_{\pm 1.3}$ & \lightG $52.8_{\pm 2.0}$ & \darkestG $56.1_{\pm 2.3}$ & $20.6_{\pm 1.3}$ & \lightG $62.7_{\pm 1.9}$ & \darkestG $63.2_{\pm 2.2}$ & \lightG $71.9_{\pm 1.4}$ & \darkestG $62.4_{\pm 1.9}$ & \darkestG $62.4_{\pm 2.2}$ \\
\quad \emph{SFT Classifier} & $32.6_{\pm 1.5}$ & $57.7_{\pm 2.0}$ & \darkestG $60.0_{\pm 2.2}$ & \lightG $74.8_{\pm 1.4}$ & $52.2_{\pm 2.0}$ & \darkestG $59.1_{\pm 2.2}$ & $14.2_{\pm 1.1}$ & \lightG $62.9_{\pm 1.9}$ & $60.6_{\pm 2.3}$ & \lightG $39.3_{\pm 1.6}$ & \darkestG $61.1_{\pm 2.0}$ & \darkestG $61.0_{\pm 2.2}$ \\
\quad \emph{Agentic Top-K} & $33.9_{\pm 1.5}$ & \darkestG $59.6_{\pm 2.0}$ & \lightG $56.9_{\pm 2.3}$ & $67.3_{\pm 1.5}$ & $42.0_{\pm 2.0}$ & \lightG $55.2_{\pm 2.2}$ & $12.3_{\pm 1.1}$ & $58.7_{\pm 2.0}$ & \darkestG $62.8_{\pm 2.2}$ & $28.2_{\pm 1.4}$ & \lightG $43.9_{\pm 2.0}$ & \lightG $53.0_{\pm 2.3}$ \\
\bottomrule[1.75pt]
\end{tabular}
}
\vspace{-0.2cm}
\end{table*}

\section{Experiment Details}
\label{sec:experiments}

% Our evaluation pipeline relies on language models to fill three distinct roles: candidate models to form teams, embedding backbones to encode text, and inference backbones for LLM-driven recruiters. 

\paragraph{Collaboration Methods}
% \label{sec:methodology:collab}

We evaluate selected team $T$ under four collaboration methods:
\emph{prompt routing}, which asks an LLM to route each query to the team member whose description best matches it; 
\emph{multi-agent refinement} \citep{du2024multiagentdebate}, in which team members iteratively critique and revise a shared draft; 
\emph{weight merging}, which fuses team members' parameters into a single model via DARE \citep{yu2023languagemodels} composed with TIES \citep{yadav2023ties}; and 
\emph{LLM blender} \citep{jiang2023llmblender}, which pairwise-ranks team members' outputs and generatively fuses the top-ranked subsets.

\paragraph{Candidate models.} We evaluate selectors across two distinct candidate pools $\mathcal{M}$. Pool 1 consists of 10 Qwen2.5-7B-Instruct~\citep{qwen2024qwen25} models trained on different instruction-tuning corpora~\citep{jiang2025sparta}. Pool 2 consists of 32 independently contributed systems from the participatory ecosystem of \citet{feng2026participation}, spanning diverse architectures, scales, and training objectives. For both pools, our goal is to select a team of $n=4$ candidate models for collaboration. Full model details are provided in Appendix~\ref{sec:appendix:pools_and_datasets}.

% \paragraph{Embedding and inference backbones.} \emph{Stated Diversity} selectors and the description component of \emph{Combined} embed descriptions uses \texttt{all-MiniLM-L6-v2}\footnote{\url{https://huggingface.co/sentence-transformers/all-MiniLM-L6-v2}}. The classifier and IRT selectors are built on \texttt{all-mpnet-base-v2}\footnote{\url{https://huggingface.co/sentence-transformers/all-mpnet-base-v2}}. Qwen2.5-7B-Instruct is used for LLM-driven selector, and the LLM-Blender ranker.

\paragraph{Datasets.} 

We evaluate collaborative performance on three main datasets: GSM8K~\citep{cobbe2021gsm8k}, TruthfulQA~\citep{lin2022truthfulqa}, and MBPP~\citep{austin2021mbpp}. Auxiliary datasets are used to construct the \emph{Performance Profile} signal and train the \emph{SFT Classifier}. \emph{Idiosyncrasies}, \emph{IRT}, and \emph{Top Solo Score} use the development set of the evaluation datasets to construct their selection signals, while all reported collaborative performance is evaluated on the test splits. Also, the \emph{SFT Classifier} is trained exclusively on teams sampled from Pool~1. When applied to Pool~2, it scores all candidate teams without further training or adaptation, constituting a cross-pool out-of-distribution transfer setting. Dataset roles, sizes, and splits are detailed in Appendix~\ref{sec:appendix:data_details}.

% All main collaboration results use three test datasets GSM8K~\citep{cobbe2021gsm8k}, TruthfulQA~\citep{lin2022truthfulqa}, and MBPP~\citep{austin2021mbpp}. \emph{Performance Profile} uses 10 independent proxy datasets with scores min--max normalized into $[\epsilon, 1]$. \emph{SFT classifier} is trained on real collaboration outcomes from three further held-out datasets. Finally, \emph{Idiosyncrasies} and \emph{IRT} selectors are trained on responses from the same test sets, but because they optimize for dispersion over abstract attribution/ability embeddings rather than maximizing accuracy, they do not directly optimize the final reported metric. 

\paragraph{Implementation.}
We use \texttt{all-MiniLM-L6-v2} to encode model descriptions and \texttt{all-mpnet-base-v2} for the learned behavioral representations. Qwen2.5-7B-Instruct serves as the backbone for LLM-based selectors and LLM-Blender. Unless otherwise specified, generation uses temperature $0.7$, top-$p$ $0.9$, and a maximum response length of 512 tokens (1024 for MBPP). Full selector and collaboration configurations are provided in Appendix~\ref{app:experimental_details}.

\paragraph{Variance estimation.}
For deterministic selectors, we estimate test-set uncertainty using $B=10{,}000$ nonparametric bootstrap resamples over item-level outcomes and report the mean with 95\% percentile intervals. For the \emph{Random} selector, we measure selection variance over five independently sampled teams and report the corresponding mean and 95\% percentile interval.

% \subsection{Hyperparameters and Evaluation}
% \label{sec:experiments:evaluation}

% \textbf{Hyperparameters Configurations.} 
% Table~\ref{tab:collab-hparams} in Appendix~\ref{app:method-configs} lists the hyperparameters used for every collaboration method $c \in \mathcal{C}$. We hold all collaboration settings fixed across selectors and pools. Decoding uses temperature of $0.7$, top-$p$ of $0.9$,  batch size of $4$, and maximum response length of $512$ tokens ($1024$ for MBPP). DARE-TIES applies uniform weighting, and LLM-Blender uses the first-listed team member as its generative fuser. Full collaboration and selector configurations are provided in Table~\ref{tab:trained-selectors} in Appendix~\ref{app:experimental-tables}.

% \textbf{Variance Estimation.} 
% For deterministic selectors, we use \emph{test set bootstrap}. Specifically, we use $B=10{,}000$ resamples (seed $0$) per (selector, method, dataset) cell, reporting the mean and 95\% percentile interval. For \emph{selection process variance}, we run the \emph{Random} selector $5$ times (seeds $0$--$4$) and report the mean and 95\% percentile interval directly over the generated teams.

% \subsection{Robustness and Scaling Study Setup}
% \label{sec:experiments:robustness}

\section{Results}
\label{sec:results}

% \lucy{what is n (team size) for all of these main results including the table? is it always the same n across selection methods?} \zw{yes the team size is fixed}

Table~\ref{tab:main_results} reports the collaborative performance for all selection strategies. The results reveal four key findings in multi-LLM team composition:

\textbf{Blind hiring masks severe variance.} 
% \lucy{e.g., interpreting this depends on the size of n proportional to the pool size} \zw{we fixed the selected team size to 4 for all the selectors}
Randomly selecting a team leads to considerable performance swings depending on the luck of the draw, yielding standard deviations that make the system unreliable. For instance, Random selection on GSM8K yields intervals of $\pm25.3$ under LLM-Blender in Pool~2 and $\pm13.4$ under DARE-TIES in Pool~1. In contrast, intentional selection strategies effectively eliminate this selection-process variance, replacing unpredictable draws with deterministic, stable teams that exhibit tight confidence intervals (typically $\pm 1.0$ to $\pm 2.5$).

\textbf{Intentional selection can build stronger teams.} 
By systematically vetting candidates, capability-aware selection strategies outperform naive heuristics. In Pool 2, simply choosing the largest models (\emph{Top Size}) yields catastrophic failures in generative and parameter-fusion settings, scoring a mere $14.4$ on GSM8K under DARE-TIES compared to \emph{Random}'s $28.4$. Conversely, capability-aware strategies can identify substantially stronger team compositions. Methods like \emph{Idiosyncrasies (least)} and \emph{Nested (most)} frequently hit the highest performance tier, outscoring the \emph{Random} baseline, the \emph{Top Solo Score} baseline, and even the \emph{Single Best Model} available in the entire pool. 

\textbf{Team selection is critical: suboptimal teams actually undermine the benefit of collaboration methods.} 
Outperforming the single strongest individual candidate remains a high, context-dependent bar. When the best solo model is exceptionally strong, assembling a team that actually surpasses it requires highly precise composition. In Pool 2, the single best model achieves $82.9$ on GSM8K. The majority of selection strategies fail to clear this threshold under Prompt Routing or DARE-TIES. However, specific optimized pairings, such as \emph{Nested (least)} under Multiagent Refine or \emph{Idiosyncrasies (least)} under LLM-Blender demonstrate that a well-composed team can still push past a highly capable solo expert. The added value of collaboration depends heavily on both the inherent difficulty of the task and the baseline competence of the available pool.

\textbf{There is no one-size-fits-all hiring strategy.} 
The success of a selection method fluctuates significantly depending on the downstream collaboration mechanism it feeds into. A strategy that recruits a state-of-the-art team for one collaborative framework can fail completely under another. For example, in Pool 1, \emph{Nested (most)} is the dominant strategy for Multiagent Refine on GSM8K ($72.9$), yet it struggles significantly when applied to MBPP under LLM-Blender ($31.6$). Conversely, \emph{Combined (most)} thrives under DARE-TIES and LLM-Blender for MBPP (scoring $59.3$ and $58.3$, respectively) but falls to $45.6$ under Multiagent Refine. This confirms that team composition cannot be decoupled from team interaction; the underlying signal used to select collaborators must be directly tailored to how the team will ultimately merge or fuse their outputs.

\section{Analysis}

\subsection{Robustness to Malicious Models}
\label{sec:results:malicious}

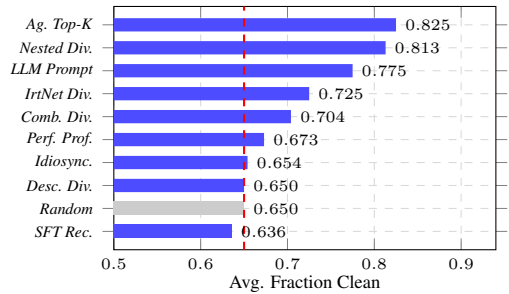
\begin{wrapfigure}{r}{0.50\textwidth}
    \centering
    \vspace{-40pt}
    \begin{tikzpicture}
        \begin{axis}[
            xbar,
            bar shift=0pt,
            width=0.95\linewidth, % Prevents labels from spilling out of the wrapfigure
            height=4.8cm, 
            xmin=0.5, xmax=0.94, % Slight padding for the right-side numbers
            ymin=0.2, ymax=10.8,
            xlabel={Avg. Fraction Clean}, % Abbreviated to save horizontal space
            xlabel style={font=\scriptsize, yshift=1ex}, 
            tick label style={font=\tiny}, 
            ytick={1,2,3,4,5,6,7,8,9,10},
            yticklabels={
                \emph{SFT Rec.},        % Heavily abbreviated
                \emph{Random},          
                \emph{Desc. Div.},      % Heavily abbreviated
                \emph{Idiosync.},       % Heavily abbreviated
                \emph{Perf. Prof.},     % Heavily abbreviated
                \emph{Comb. Div.},      % Heavily abbreviated
                \emph{IrtNet Div.},     % Heavily abbreviated
                \emph{LLM Prompt},
                \emph{Nested Div.},     % Heavily abbreviated
                \emph{Ag. Top-K}        % Heavily abbreviated
            },
            nodes near coords={\pgfmathprintnumber[fixed, zerofill, precision=3]\pgfplotspointmeta}, % Added zerofill to force 3 decimal places
            nodes near coords align={horizontal},
            nodes near coords style={font=\tiny, text=black}, 
            point meta=x,
            grid=major,
            grid style={dashed, gray!30}, 
            bar width=5pt, % Slimmed down further
        ]
        
        % 1. SFT Classifier (Original Blue)
        \addplot[fill=blue!70!white, draw=none] coordinates {(0.636, 1)};
        
        % 2. Random Baseline (Original Gray)
        \addplot[fill=gray!40!white, draw=none] coordinates {(0.650, 2)};
        
        % 3-10. The rest of the methods (Original Blue)
        \addplot[fill=blue!70!white, draw=none] coordinates {
            (0.650, 3)
            (0.654, 4)
            (0.673, 5)
            (0.704, 6)
            (0.725, 7)
            (0.775, 8)
            (0.813, 9)
            (0.825, 10)
        };
        
        % Vertical line (Original Red) - Marks the Random Baseline threshold (0.650)
        \draw[red, thick, dashed] (axis cs:0.650,\pgfkeysvalueof{/pgfplots/ymin}) -- (axis cs:0.650,\pgfkeysvalueof{/pgfplots/ymax});

        \end{axis}
    \end{tikzpicture}
    \vspace{-10pt}
    \caption{Average fraction of clean models in selected teams across contamination levels and randomized injection trials. The red dashed line marks the Random baseline.}
    \label{fig:malicious_ranking}
    \vspace{-10pt} 
\end{wrapfigure}

To evaluate robustness to misaligned candidates, we introduce six misaligned models from \citet{yang2026amongus}. Their descriptions follow the same innocuous one-line format as aligned models, providing no explicit signal of misalignment to text-based selectors. At each contamination level $r$, we inject $r$ misaligned models into a pool of 10 models and repeat the experiment 10 times with randomized removal and injection orders. We measure the fraction of aligned models in the selected four-model team with an expected fraction of $(10-r)/10$ under random selection.

Figure~\ref{fig:malicious_ranking} reveals a contrast between methods that passively rely on indirect signals and those that explicitly evaluate model capability. Unanchored diversity-based selectors are particularly vulnerable. Semantic selectors like \emph{Description Diversity} perform near a random baseline when model descriptions are misleading. A full round-by-round breakdown can be found in Appendix~\ref{app:malicious_grid}.

In contrast, some selection methods equipped with explicit capability filters or agentic evaluation outperform the random baseline and are better equipped to avoid compromised candidates. \emph{Agentic Top-K} achieves the highest clean-model fraction, consistent with its interactive interview process providing a stronger capability signal. Similarly, \emph{Nested} first filters out lower-capability candidates before applying diversity-based selection, reducing the exposure of the downstream diversity search to malicious candidates. Overall, these results suggest that robustness benefits from selection in observed model behavior or capability, rather than relying solely on stated descriptions.

% \emph{Agentic Top-K}'s interactive interview process helps probe the models and expose their flaws. Similarly, \emph{Nested Diverse} tends to protect the final team by filtering out lower-capability models initially, isolating the downstream diversity search from poisoned candidates. Both mechanisms reduce the influence of compromised models, suggesting that robustness depends on grounding selection in observed capability rather than relying solely on description diversity, or learned proxies.

\subsection{Effects of Pool-Size Scaling}
\label{sec:results:scaling}
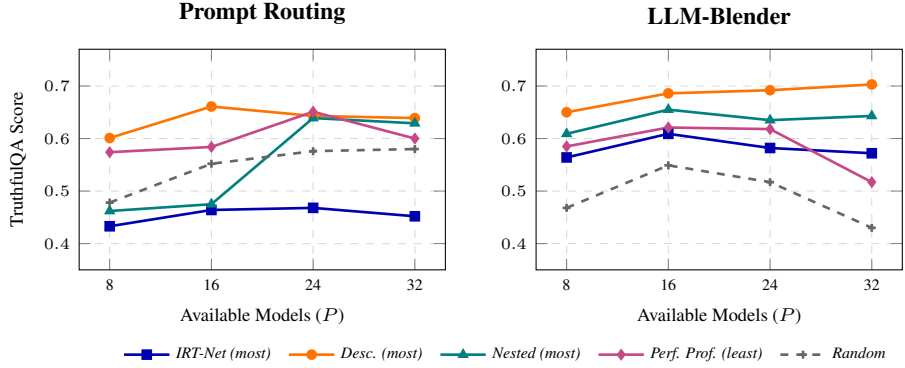
\begin{figure}[t]
    \centering
    \vspace{-10pt}
    \begin{tikzpicture}
        \begin{groupplot}[
            group style={
                group size=2 by 1,
                horizontal sep=1.2cm
            },
            width=0.46\textwidth,
            height=4.5cm,
            xlabel={Available Models ($P$)},
            label style={font=\scriptsize},
            tick label style={font=\tiny},
            symbolic x coords={8, 16, 24, 32},
            xtick=data,
            ymin=0.35, ymax=0.77,
            ytick={0.4, 0.5, 0.6, 0.7},
            grid=major,
            grid style={dashed, gray!30},
            % thick,
            every axis plot/.append style={
                line width=1pt,
                mark size=1.5pt
            }
        ]

        % ==========================================
        % SUBPLOT 1: Prompt Routing
        % ==========================================
        \nextgroupplot[
            title={\footnotesize \textbf{Prompt Routing}},
            ylabel={TruthfulQA Score},
            ylabel style={font=\scriptsize}
        ]

        \addplot[color=blue!70!black, mark=square*] coordinates {
            (8, 0.433) (16, 0.464) (24, 0.468) (32, 0.452)
        };

        \addplot[color=orange!90!red, mark=*] coordinates {
            (8, 0.601) (16, 0.661) (24, 0.643) (32, 0.639)
        };

        \addplot[color=teal, mark=triangle*] coordinates {
            (8, 0.462) (16, 0.475) (24, 0.639) (32, 0.629)
        };

        \addplot[color=magenta!80!black, mark=diamond*] coordinates {
            (8, 0.574) (16, 0.584) (24, 0.651) (32, 0.600)
        };

        \addplot[color=gray!80!black, mark=+, dashed] coordinates {
            (8, 0.478) (16, 0.552) (24, 0.576) (32, 0.580)
        };

        % ==========================================
        % SUBPLOT 2: LLM-Blender
        % ==========================================
        \nextgroupplot[
            title={\footnotesize \textbf{LLM-Blender}},
            legend style={
                at={(-0.08,-0.30)},
                anchor=north,
                legend columns=5,
                draw=none,
                fill=none,
                font=\tiny,
                /tikz/every even column/.append style={
                    column sep=0.15cm
                }
            }
        ]

        \addplot[color=blue!70!black, mark=square*] coordinates {
            (8, 0.564) (16, 0.609) (24, 0.582) (32, 0.572)
        };
        \addlegendentry{\emph{IRT-Net (most)}}

        \addplot[color=orange!90!red, mark=*] coordinates {
            (8, 0.650) (16, 0.686) (24, 0.692) (32, 0.703)
        };
        \addlegendentry{\emph{Desc. (most)}}

        \addplot[color=teal, mark=triangle*] coordinates {
            (8, 0.609) (16, 0.655) (24, 0.635) (32, 0.643)
        };
        \addlegendentry{\emph{Nested (most)}}

        \addplot[color=magenta!80!black, mark=diamond*] coordinates {
            (8, 0.585) (16, 0.621) (24, 0.618) (32, 0.517)
        };
        \addlegendentry{\emph{Perf. Prof. (least)}}

        \addplot[color=gray!80!black, mark=+, dashed] coordinates {
            (8, 0.468) (16, 0.549) (24, 0.517) (32, 0.430)
        };
        \addlegendentry{\emph{Random}}

        \end{groupplot}
    \end{tikzpicture}

    \caption{\textbf{Pool-size scaling on Pool 2.} Expanding the uncurated pool without strict capability constraints causes team performance to saturate or even degrade.}
    %\vspace{-10pt}
    \label{fig:pool_scaling}
\end{figure}

To evaluate how selection strategies scale with candidate pool size, we vary the number of available models in Pool~2 as $P \in \{8, 16, 24, 32\}$ while fixing the selected team size to $n=4$. This setting allows us to examine how different selectors behave as the number of candidate collaborators increases. Figure~\ref{fig:pool_scaling} shows how collaborative performance changes as the candidate pool expands. Initially, increasing the pool from small ($P=8$) to moderately large ($P=16$) helps some selection methods build stronger teams, suggesting that additional candidates provide useful opportunities for identifying complementary models. However, further expanding the pool ($P=32$) yields diminishing or negative returns for several strategies. Semantic diversity methods tend to plateau,  while some unconstrained approaches degrade at larger pool sizes, particularly LLM-Blender. 

The pronounced drop of the \emph{Random} baseline at $P=32$ is consistent with the expanded pool containing substantially weaker candidates.
% The pronounced collapse of the random baseline at $P=32$ suggests that expanding the pool can include a ``long tail'' of substantially degraded models. 
Consequently, methods that prioritize diversity without accounting for baseline capability become susceptible to selecting these weak outliers into the team. Overall, these results suggest that simply scaling up the candidate pool is not sufficient to improve collaboration. As the pool expands, selection must balance diversity with candidate capability to avoid incorporating weak but superficially distinctive models.
% Overall, these results suggest that simply scaling up the candidate pool is not sufficient to improve collaboration: as the pool expands, capability-aware filtering becomes increasingly important for distinguishing useful diversity from low-quality variation and preventing collaborative performance from deteriorating.

\subsection{Effects of Pool Diversity Scaling}
\label{sec:results:saturation}

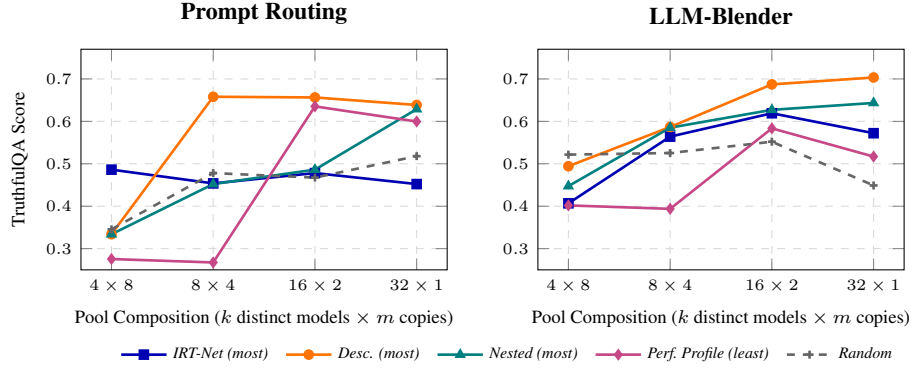
\begin{figure}[t]
    \vspace{-10pt}
    \centering
    \begin{tikzpicture}
        \begin{groupplot}[
            group style={
                group size=2 by 1,
                horizontal sep=1.2cm
            },
            width=0.46\textwidth,
            height=4.5cm,
            xlabel={Pool Composition ($k$ distinct models $\times$ $m$ copies)},
            label style={font=\scriptsize},
            tick label style={font=\tiny},
            symbolic x coords={4x8, 8x4, 16x2, 32x1},
            xticklabels={$4\times 8$, $8\times 4$, $16\times 2$, $32\times 1$},
            xtick=data,
            ymin=0.25, ymax=0.77,
            ytick={0.3, 0.4, 0.5, 0.6, 0.7},
            grid=major,
            grid style={dashed, gray!30},
            % thick,
            every axis plot/.append style={
                line width=1pt,
                mark size=1.5pt
            }
        ]

        % ==========================================
        % SUBPLOT 1: Prompt Routing
        % ==========================================
        \nextgroupplot[
            title={\footnotesize \textbf{Prompt Routing}},
            ylabel={TruthfulQA Score},
            ylabel style={font=\scriptsize}
        ]

        \addplot[color=blue!70!black, mark=square*] coordinates {
            (4x8, 0.4862) (8x4, 0.4538) (16x2, 0.4781) (32x1, 0.4522)
        };

        \addplot[color=orange!90!red, mark=*] coordinates {
            (4x8, 0.3339) (8x4, 0.6580) (16x2, 0.6564) (32x1, 0.6386)
        };

        \addplot[color=teal, mark=triangle*] coordinates {
            (4x8, 0.3339) (8x4, 0.4522) (16x2, 0.4862) (32x1, 0.6288)
        };

        \addplot[color=magenta!80!black, mark=diamond*] coordinates {
            (4x8, 0.2755) (8x4, 0.2674) (16x2, 0.6353) (32x1, 0.5997)
        };

        \addplot[color=gray!80!black, mark=+, dashed] coordinates {
            (4x8, 0.3455) (8x4, 0.4781) (16x2, 0.4674) (32x1, 0.5180)
        };

        % ==========================================
        % SUBPLOT 2: LLM-Blender
        % ==========================================
        \nextgroupplot[
            title={\footnotesize \textbf{LLM-Blender}},
            legend style={
                at={(-0.08,-0.30)},
                anchor=north,
                legend columns=5,
                draw=none,
                fill=none,
                font=\tiny,
                /tikz/every even column/.append style={column sep=0.15cm}
            }
        ]

        \addplot[color=blue!70!black, mark=square*] coordinates {
            (4x8, 0.4068) (8x4, 0.5640) (16x2, 0.6191) (32x1, 0.5721)
        };
        \addlegendentry{\emph{IRT-Net (most)}}

        \addplot[color=orange!90!red, mark=*] coordinates {
            (4x8, 0.4943) (8x4, 0.5867) (16x2, 0.6872) (32x1, 0.7034)
        };
        \addlegendentry{\emph{Desc. (most)}}

        \addplot[color=teal, mark=triangle*] coordinates {
            (4x8, 0.4473) (8x4, 0.5851) (16x2, 0.6272) (32x1, 0.6434)
        };
        \addlegendentry{\emph{Nested (most)}}

        \addplot[color=magenta!80!black, mark=diamond*] coordinates {
            (4x8, 0.4019) (8x4, 0.3938) (16x2, 0.5835) (32x1, 0.5170)
        };
        \addlegendentry{\emph{Perf.\ Profile (least)}}

        \addplot[color=gray!80!black, mark=+, dashed] coordinates {
            (4x8, 0.5216) (8x4, 0.5254) (16x2, 0.5520) (32x1, 0.4490)
        };
        \addlegendentry{\emph{Random}}

        \end{groupplot}
    \end{tikzpicture}

    \caption{\textbf{Pool diversity scaling on Pool 2.} Evaluating a 32-slot pool across varying compositions of $k$ distinct models $\times$ $m$ copies. Naive diversity (\emph{Desc.}) saturates as $k$ increases, while behavioral diversity (\emph{IRT-net}) degrades at $32\times 1$, risking the inclusion of low-quality outliers.}
    %\vspace{-10pt}
    \label{fig:saturation_subplots}
\end{figure}

To isolate the effect of candidate diversity from pool size, we fix the pool at 32 candidate slots and the selected team size at $n=4$, while varying the number of distinct models $k$ and the number of copies per model $m$: $(k,m)\in\{(4,8),(8,4),(16,2),(32,1)\}$. Here, we operationalize pool diversity by the number of distinct model identities while holding the total number of candidate slots fixed. We evaluate downstream performance on TruthfulQA. Figure~\ref{fig:saturation_subplots} shows how collaborative performance changes as the number of distinct candidates increases.

For several selectors, replacing duplicate candidates with distinct models initially improves performance, but the gains tend to saturate as $k$ increases. At the highest-diversity setting ($32\times1$), some selectors even degrade, indicating that greater distinctness does not necessarily translate into more useful team composition. Once the pool already covers a broad range of capabilities, additional distinct models may instead introduce weaker candidates. Overall, these results suggest that reducing redundancy can benefit collaboration, but maximizing distinctness alone is insufficient; effective selection must balance candidate diversity with capability.

\begin{wraptable}{R}{0.5\textwidth}
\centering
\caption{\textbf{OOD evaluation on Pool 2.} The highest mean score in each column per pool is bolded.}
\label{tab:ood_results}
\scriptsize
\setlength{\tabcolsep}{1pt}
\resizebox{\linewidth}{!}{%
\begin{tabular}{@{} l c c c c @{}}
\toprule[1.75pt]
& \multicolumn{2}{c}{CoCoNot} & \multicolumn{2}{c}{GPQA-Diamond} \\
\cmidrule(lr){2-3} \cmidrule(l){4-5}
\textbf{Selector} & \textbf{Routing} & \textbf{Blender} & \textbf{Routing} & \textbf{Blender} \\
\midrule
Random & $44.3_{\pm 6.9}$ & $26.1_{\pm 9.2}$ & $22.4_{\pm 2.3}$ & $26.9_{\pm 11.0}$ \\
IRT-Net (most) & $38.6_{\pm 3.0}$ & $29.7_{\pm 3.0}$ & $19.2_{\pm 7.5}$ & $33.3_{\pm 9.0}$ \\
Idio. (most) & $45.3_{\pm 3.0}$ & $25.1_{\pm 3.0}$ & $17.2_{\pm 7.5}$ & $\mathbf{40.4}_{\pm 9.5}$ \\
SFT & $\mathbf{48.6}_{\pm 3.0}$ & $\mathbf{33.1}_{\pm 3.0}$ & $\mathbf{28.3}_{\pm 9.0}$ & $34.3_{\pm 9.0}$ \\
\bottomrule[1.75pt]
\end{tabular}
}
\vspace{-1em}
\end{wraptable}

\subsection{Does Capability-Aware Selection Generalize Out-of-Distribution?}
\label{sec:analysis:ood}

The behavioral and learned selectors in Table~\ref{tab:main_results} construct their selection signals from task-specific or auxiliary data. To examine whether these signals transfer to unseen task domains, we evaluate selected methods on CoCoNot (safety) and GPQA-Diamond (graduate-level science). we evaluate three learned behavioral selectors---\emph{IRT-Net}, \emph{Idiosyncrasies}, and \emph{SFT Classifier}---on two unseen task domains. \emph{Performance Profile} is excluded from this analysis because their construction suite includes GPQA-Diamond and CoCoNot (Table~\ref{tab:ood_results}).

The \emph{SFT Classifier} transfers consistently across the two unseen domains, outperforming \emph{Random} in all four settings and achieving the highest mean performance in three of them. In contrast, representation-based behavioral signals show more scenario-dependent transfer: \emph{IRT-Net} and \emph{Idiosyncrasies} underperform \emph{Random} in some Prompt Routing settings, while both improve over \emph{Random} on GPQA-Diamond under LLM-Blender.

% The classifier-based selectors in Table~\ref{tab:main_results} were evaluated in-distribution. To test their generalization, we evaluate all classifier-based selection methods on two novel, out-of-distribution (OOD) task types relative to their training data: CoCoNot (safety) and GPQA-Diamond (graduate level science). For each capability-aware signal, we evaluate the most variant. These pre-selected Pool 2 teams are evaluated using \textsc{prompt routing} and \textsc{LLM-Blender} (Table~\ref{tab:ood_results}).

% The OOD results show that supervised team-level selection generalizes robustly across domains, while representation-based diversity signals show scenario-dependent transfer. Specifically, the \emph{SFT Classifier} consistently outperforms the \emph{Random} baseline across all four evaluated settings, achieving the highest performance in three of them. In contrast, unsupervised diversity signals (\emph{IRT} and \emph{Idiosyncrasies}) struggle under \textsc{prompt routing}, but show strong performance when paired with \textsc{LLM-Blender} where \emph{Idiosyncrasies} achieves the overall top score on GPQA-Diamond.

\section{Conclusion}

In this work, we formalized the problem of collaborator selection in multi-LLM systems and introduced a comprehensive taxonomy of selection strategies. Through extensive evaluation across diverse candidate pools, tasks, and collaboration mechanisms, we demonstrated that the composition of an LLM team is just as critical as the method by which they collaborate. Relying on blind random sampling or simplistic heuristics introduces severe variance and limits the potential of multi-agent frameworks. Our findings highlight the importance of grounding selection in observed model capabilities rather than relying solely on model descriptions or simple heuristics. These principled strategies successfully scale with pool size, resist the inclusion of misaligned models, and generalize to novel task distributions. Ultimately, as the ecosystem of open-source models continues to expand, building effective collaborative AI systems will require shifting our focus from merely \emph{how} models interact to \emph{who} is hired for the job in the first place.

\subsection*{AI use statement}

In this work, we used a generative AI tool to assist
with drafting and editing manuscript text, and to help
identify and verify citations for prior work discussed throughout the paper. We did not use generative AI tools to design the selection algorithms, collaboration
methods, or experimental protocols presented in this work, nor to produce the underlying experimental results: all reported scores were obtained by running the described selection methods and MoCo collaboration methods on the stated benchmarks, independent of any AI tool.

All AI-assisted text was reviewed and edited by the authors, and all factual and numerical claims traceable to the underlying experimental data were checked against that data by the authors. We take responsibility for the final content of this work, including all text, claims, and artifacts produced with the aid of generative AI.

\subsection*{Ethics statement}

This work does not involve human subjects; the ``interview'' procedure in Section~\ref{app:agentic-interview} is an automated exchange between two language models and involves no human participants, data, or annotation.

The adversarial-model-detection study deliberately incorporates six models fine-tuned to be misaligned, drawn from the existing, published Among Us benchmark rather than newly created for this paper. These models are used strictly to test whether selection strategies can identify and exclude them from a collaborating team; they are not released, deployed, or used for any purpose beyond this detection benchmark, and no new misaligned model is introduced by this work.

All base models and benchmarks used in this work are obtained from publicly released, appropriately licensed sources, and we do not introduce new data collected from or about individuals. We are not aware of a direct harmful application of this work's findings.

\subsection*{Reproducibility Statement}
% This work is fully reproducible. All code, datasets, and experiment logs required to replicate the reported results duanm@usc.edu  be released at \zw{add anonymized link}.
All code, datasets, and experiment logs required to reproduce the reported results will be released at a repository upon publication.

\bibliography{iclr2027_conference}
\bibliographystyle{iclr2027_conference}

\appendix

\section{Limitations}
\label{sec:limitations}
While principled team selection improves multi-LLM systems, several limitations remain.

\textbf{Computational overhead.} Exact dispersion search scales combinatorially ($\binom{P}{n}$), becoming intractable for large pools or teams. While greedy approximations and LLM-based recruiters accelerate inference, their upfront signal-generation costs remain high: IRT and idiosyncrasies classifiers require evaluating every candidate on full proxy datasets, and \textsc{Agentic top-$k$} requires multi-turn interviews per member.

\textbf{Decoupled selection and collaboration.} We treat downstream collaboration mechanisms as fixed black boxes using default configurations. We do not jointly optimize the selection algorithm and the collaboration mechanism (e.g., co-training the SFT Classifier alongside DARE-TIES scaling weights).

\textbf{Model scale.} We evaluate open-weight models up to 14B parameters. Collaboration dynamics may shift significantly with frontier-class models (70B+ or closed APIs), which might exhibit less behavioral dispersion or unlock novel synergistic benefits during refinement that smaller pools cannot capture.

\section{Design Choice: Max-Min vs. Average Dispersion}
\label{app:maxmin-vs-avg}

\paragraph{Algorithmic Difference.} Algorithm~\ref{alg:exact} prioritizes max-min dispersion: it evaluates a team by its maximum pairwise similarity and breaks ties using average pairwise similarity. Conversely, average dispersion optimizes $\mathrm{avgSim}$ as the primary objective and uses $\mathrm{maxSim}$ as the tie-breaker.

\paragraph{Performance Comparison.} In a head-to-head evaluation across 16 settings (Table~\ref{tab:maxmin-vs-avg}), 7 generated identical teams where minor score variations resulted solely from generation sampling noise. Across the 9 settings where the objectives selected different teams, average dispersion won slightly more frequently (5 vs.\ 4) by small margins ($\le 0.101$). However, max-min achieved the two largest performance gains. Thus, max-min trades negligible typical-case margins for vital worst-case protection against severe performance collapses in redundant candidate pools.

\paragraph{Rationale for Max-Min.} We select max-min primary dispersion because a team's functional complementarity is bottlenecked by its most redundant pair; optimizing average similarity allows near-duplicate models as long as other pairs are highly distinct. Max-min acts as a robust worst-case safeguard against severe performance degradation, while using average similarity as a tie-breaker preserves overall team structure.

\begin{algorithm}[H]
\caption{Exact Max-Min Diversity Search (Algorithm~\ref{alg:exact}, repeated for comparison)}
\begin{algorithmic}[1]
\REQUIRE $\mathrm{sim} \in \mathbb{R}^{P\times P}$, team size $n$, mode $\in \{\textsc{most}, \textsc{least}\}$
\STATE $T^\star \gets \varnothing$
\FOR{each $T \subseteq \{1,\dots,P\}$ with $|T| = n$}
    \STATE $\mathrm{\textbf{maxSim}}(T) \gets \max_{i,j \in T} \mathrm{sim}(i,j)$
    \STATE $\mathrm{avgSim}(T) \gets \mathrm{mean}_{i,j \in T}\, \mathrm{sim}(i,j)$
    \IF{mode $= \textsc{most}$ \textbf{and} $(\mathbf{\mathrm{maxSim}(T)}, \mathrm{avgSim}(T))$ lex.\ smaller than incumbent}
        \STATE $T^\star \gets T$
    \ELSIF{mode $= \textsc{least}$ \textbf{and} $(\mathbf{\mathrm{maxSim}(T)}, \mathrm{avgSim}(T))$ lex.\ larger than incumbent}
        \STATE $T^\star \gets T$
    \ENDIF
\ENDFOR
\RETURN $T^\star$
\end{algorithmic}
\end{algorithm}

\vspace{1em}

\begin{algorithm}[H]
\caption{Exact Average-Similarity Diversity Search}
\begin{algorithmic}[1]
\REQUIRE $\mathrm{sim} \in \mathbb{R}^{P\times P}$, team size $n$, mode $\in \{\textsc{most}, \textsc{least}\}$
\STATE $T^\star \gets \varnothing$
\FOR{each $T \subseteq \{1,\dots,P\}$ with $|T| = n$}
    \STATE $\mathrm{\textbf{avgSim}}(T) \gets \mathrm{mean}_{i,j \in T}\, \mathrm{sim}(i,j)$
    \STATE $\mathrm{maxSim}(T) \gets \max_{i,j \in T} \mathrm{sim}(i,j)$
    \IF{mode $= \textsc{most}$ \textbf{and} $(\mathbf{\mathrm{avgSim}(T)}, \mathrm{maxSim}(T))$ lex.\ smaller than incumbent}
        \STATE $T^\star \gets T$
    \ELSIF{mode $= \textsc{least}$ \textbf{and} $(\mathbf{\mathrm{avgSim}(T)}, \mathrm{maxSim}(T))$ lex.\ larger than incumbent}
        \STATE $T^\star \gets T$
    \ENDIF
\ENDFOR
\RETURN $T^\star$
\end{algorithmic}
\end{algorithm}

\begin{table}[htbp]
\centering
\caption{Max-min vs.\ average dispersion score comparison. Tasks are evaluated using LLM-Blender on TruthfulQA. Higher scores per setting are in \textbf{bold}.}
\label{tab:maxmin-vs-avg}
\small
\setlength{\tabcolsep}{5pt}
\begin{tabular}{@{} l l c c c c @{}}
\toprule[1.75pt]
& & \multicolumn{2}{c}{\textbf{Pool 1}} & \multicolumn{2}{c}{\textbf{Pool 2}} \\
\cmidrule(lr){3-4} \cmidrule(l){5-6}
\textbf{Category} & \textbf{Direction} & \textbf{Max-Min} & \textbf{Average} & \textbf{Max-Min} & \textbf{Average} \\
\midrule
Stated Diversity & Most  & \textbf{0.621} & 0.606 & \textbf{0.703} & 0.697 \\
                 & Least & \textbf{0.491} & 0.488 & \textbf{0.559} & 0.332 \\
\midrule
Combined         & Most  & 0.569 & \textbf{0.592} & 0.336 & \textbf{0.428} \\
                 & Least & \textbf{0.584} & 0.574 & \textbf{0.530} & 0.209 \\
\midrule
Nested           & Most  & 0.551 & \textbf{0.580} & \textbf{0.643} & 0.541 \\
                 & Least & 0.551 & \textbf{0.575} & \textbf{0.501} & 0.499 \\
\midrule
Perf. Profile    & Most  & \textbf{0.506} & 0.494 & 0.452 & \textbf{0.514} \\
                 & Least & \textbf{0.587} & 0.571 & 0.517 & \textbf{0.618} \\
\bottomrule[1.75pt]
\end{tabular}
\end{table}

\section{Additional Selection Method Details}
\label{app:learned-selectors}

\paragraph{LLM Prompt.}
The LLM-prompt recruiter directly selects a team from candidate model descriptions. Given the descriptions $\mathcal{Z}=\{\operatorname{desc}(m_i)\}_{i=1}^{P}$ and the target team size $n$, we prompt an LLM recruiter with the full candidate pool and ask it to select $n$ models that would form an effective collaborative team. The recruiter is instructed to consider both individual capabilities and potential complementarity among team members, and returns the selected model identifiers directly. We provide the full prompt in Appendix~\ref{app:prompt}.

\paragraph{SFT Classifier.}
The SFT Classifier predicts the collaborative performance of a candidate team directly from its member descriptions. Given a candidate team $T=\{m_{i_1},\ldots,m_{i_n}\}$, we concatenate the descriptions of its members and use a learned scorer $f_{\phi}$ to produce a predicted team score,
\[
    \hat{y}_T =
    f_{\phi}\big(
    \operatorname{desc}(m_{i_1}),\ldots,
    \operatorname{desc}(m_{i_n})
    \big).
\]
Training targets are obtained by executing sampled teams under the collaboration methods and recording their downstream performance. At inference time, the scorer evaluates each candidate team of size $n$, and teams are ranked by their predicted scores for selection.

\paragraph{Agentic top-$k$.}
The agentic recruiter evaluates candidates individually through adaptive multi-turn interviews. For each candidate $m_i$, an LLM interviewer iteratively selects a capability axis and generates a question conditioned on the interview history. After turns, the complete transcript is evaluated to obtain an initial interview score. Because independently assigned scores tend to be compressed, we subsequently perform comparative re-scoring by presenting multiple anonymized candidate transcripts jointly to the interviewer. This produces the selection signal
\[
    \mathcal{Z}=\{q_i\}_{i=1}^{P},
\]
where $q_i$ denotes the calibrated interview score of candidate $m_i$. Candidates are then ranked by $q_i$, and the top $n$ candidates form the selected team. The full interview and scoring protocol is provided in Appendix~\ref{app:agentic-interview}.

\section{Additional Candidate Pools Details}
\label{sec:appendix:pools_and_datasets}

In our experiment, Pool~1 contains 10 Qwen2.5-7B models that share the same backbone but differ in their fine-tuning corpora, providing a controlled setting in which model variation primarily reflects post-training data. Pool~2 contains 32 models spanning diverse source architectures, scales, training objectives, and corpora, providing a substantially more heterogeneous candidate pool.

Table~\ref{tab:pool1-models} lists all 10 Pool~1 models; Table~\ref{tab:pool2-models} lists all 32 Pool~2 models. For Pool~2, size is the parameter count of the \emph{original} contributed model each checkpoint was distilled from and it is used by the \emph{Top-size} baseline. Every deployed Pool-2 checkpoint is itself a uniform Qwen2.5-7B distillation regardless of this value. Two sizes (\texttt{parti\_6}, \texttt{parti\_21}) and two base architectures (\texttt{parti\_6}, \texttt{parti\_15}) were not stated in the original repository name and were confirmed against the corresponding Hugging Face model card; where a base architecture could not be confirmed from either source, we mark it ``unspecified'' rather than assume one. ``ID'' links the deployed distilled checkpoint; ``Original source'' links the pre-distillation contributed model.

\section{Additional Data Details}
\label{sec:appendix:data_details}

% \subsection{Evaluation Metrics}
% \label{sec:appendix:metrics}
% We evaluate collaborative configurations primarily through their absolute task performance, denoted as $\textit{score}(T,c,d)$ for a given team $T$, collaboration mechanism $c$, and dataset $d$. Reported scores isolate two sources of variance. For deterministic selectors, we estimate test-set variance via nonparametric bootstrap resampling over item-level outcomes. For the \emph{Random} baseline, we measure selection variance directly across independently seeded team draws. 
% Exact estimation parameters are detailed in Section~\ref{sec:experiments:evaluation}.

We use four groups of datasets for distinct experimental purposes: main collaboration evaluation, training the SFT Classifier, constructing model performance profiles, and out-of-distribution (OOD) evaluation. Tables~\ref{tab:main-datasets}, \ref{tab:sft-datasets}, \ref{tab:perf-profile-datasets}, and \ref{tab:ood-datasets} summarize the datasets used for each role.

The main collaboration datasets are summarized in Table~\ref{tab:main-datasets}. Each dataset is split into disjoint development and test sets, with all final collaborative performance reported on the test set. The development set is used only to construct selection signals for methods that require task-specific behavioral information. Specifically, \emph{Top solo score} ranks candidate models by their individual performance on the development set, while \emph{Idiosyncrasies} and \emph{IRT} construct their behavioral signals from model responses on the development set. The selected teams are then evaluated exclusively on the held-out test set. None of the selectors evaluated in the OOD experiment uses GPQA-Diamond or CoCoNot to construct its selection signal.

The three main evaluation datasets are disjoint from the auxiliary datasets used for SFT training and performance profiling. For training the SFT-Recruiter, we use BBH, MMLU-Redux, and HumanEval. For constructing \emph{Performance Profile}, we use a separate suite of held-out benchmarks: ARC-Challenge, BBH, MMLU-Redux, GPQA-Diamond, MATH, MedQA, PopQA, HumanEval, CoCoNot, and Human-Interest. For OOD evaluation, we use GPQA-Diamond and CoCoNot.

% We use three groups of datasets for distinct experimental purposes: the main collaboration evaluation, training the SFT-based team scorer, constructing model performance profiles and out-of-distribution evaluation. Tables~\ref{tab:main-datasets}, \ref{tab:sft-datasets}, \ref{tab:perf-profile-datasets} and \ref{tab:ood-datasets} summarize the datasets used for each role.

% The main collaboration evaluation is in Tables~\ref{tab:main-datasets}. It have dev set and test set and we report final results in test set. We use dev set to select best performing models in Top solo score and report score in testset. 

% The three main evaluation datasets are disjoint from the auxiliary datasets used for SFT training and performance profiling. 

% SFT training use BBH, MMLU-Redux, and HumanEval. 

% Performance-profile use the ARC-Challenge, BBH, MMLU-redux, GPQA-Diamond, MATH, MedQA, PopQA, HumanEval, CocoNot, and Human-interest.

% Some auxiliary datasets serve multiple roles: BBH and MMLU-Redux are used in full for performance profiling and as separate 150-example subsets for SFT training; HumanEval is used for both purposes; and GPQA-Diamond and CoCoNot additionally serve as the OOD evaluation sets in Section~\ref{sec:analysis:ood}. Except for the explicitly documented Idiosyncrasies Classifier and IRT settings (Section~\ref{sec:experiments}), model-selection signals do not directly use the three main evaluation datasets.

\section{Additional Experimental Details}
\label{app:experimental_details}

\subsection{Collaboration Method Configurations}
\label{app:method-configs}

For \emph{Prompt routing}, each query is routed according to the descriptions of the team members. \emph{Multi-agent refine} performs three refinement rounds. For \emph{Weight merging}, we use DARE-TIES with Qwen2.5-7B-Instruct as the backbone and \emph{average} as the merging mode. For \emph{LLM-Blender}, we use Qwen2.5-7B-Instruct as the backbone, set top-$k$ to $3$, and use the first team member as the fuser.

% \subsection{Selection Method Configurations}
% \label{app}

\subsection{Selection Method Configurations}
\label{app:selection-configs}

\paragraph{Idiosyncrasies.}

We use \texttt{all-mpnet-base-v2} with full fine-tuning, mean pooling, and a linear classification head. We train the classifier for 5 epochs with batch size 32, learning rate $2\times10^{-4}$, AdamW, and a cosine learning-rate schedule. We fit the Idiosyncrasies classifier using item-level outcomes from the development split of each main evaluation dataset.
% For each model, responses are split 90/10 into training and validation sets, and the confusion matrix is constructed using only held-out validation responses.

\paragraph{IRT Ability.}
We learn a 232-dimensional ability embedding for each model using a 12-expert dense MoE ($4\times$ the number of datasets) over a frozen 768-dimensional query encoder. The model is trained for 50 epochs with batch size 256, learning rate $10^{-3}$, Adam, a cosine learning-rate schedule, binary cross-entropy loss, and embedding weight decay of $0.1$. We fit the IRT model using item-level success/failure outcomes from the development split of each main evaluation dataset.

\paragraph{SFT Classifier.}
We fine-tune \texttt{Qwen2.5-7B-Instruct} with LoRA ($r=16$, $\alpha=32$, dropout $0.05$, targeting the $\{q,k,v,o\}$ projections) and a sequence-classification head with sigmoid output. 
% The scorer is trained on 168 team--method examples from 42 sampled four-model teams evaluated under four collaboration methods. 
We train for 40 epochs with batch size 4, learning rate $10^{-4}$, and MSE loss, using 6-fold group cross-validation by team identity. At inference time, the scorer evaluates all $\binom{10}{4}=210$ candidate teams in Pool~1. For Pool~2, it evaluates all $\binom{32}{4}=35{,}960$ teams, constituting an out-of-distribution transfer setting relative to Pool~1.

\paragraph{Agentic top-$k$.}
We use \texttt{Qwen2.5-7B-Instruct} as the interviewer. Each candidate undergoes a four-turn adaptive interview covering five axes: reasoning, code, factual knowledge, calibrated honesty, and open-ended conversational quality. Interviewer questions and verdicts use greedy decoding, while candidate responses are sampled with temperature $0.7$ and top-$p$ $0.9$. Initial independent ratings on a 1--10 scale are followed by comparative re-scoring using anonymized batches of up to 10 candidates. For Pool~2, whose 32 candidates exceed the single-batch limit, scores are averaged over three independently re-batched comparative rounds, with residual ties resolved by a final strict-ranking pass. Full interview prompts and scoring details are provided in Appendix~\ref{app:agentic-interview}.

\subsection{Compute and Implementation}
\label{app:compute}

All experiments run on a mixture of NVIDIA A40, A100, and L40S GPUs with bfloat16 precision. Collaboration methods execute through MoCo~\citep{feng2026moco}, installed unmodified from its published release and used strictly as a black box. Selection methods are implemented in PyTorch and HuggingFace Transformers, utilizing PEFT for LoRA adapters and the \texttt{sentence-transformers} library for embedding backbones.

\begin{table}[htbp]
\centering
\caption{Main evaluation datasets}
\vspace{1mm}
\label{tab:main-datasets}
\small
\begin{tabular}{@{} l l l l @{}}
\toprule
Dataset & Task type & Dev set size & Test set size \\
\midrule
GSM8K \citep{cobbe2021gsm8k} & exact match & 200 & 1000 \\
TruthfulQA \citep{lin2022truthfulqa} & multiple choice & 200 & 617 \\
MBPP \citep{austin2021mbpp} & coding & 487 & 487 \\
\bottomrule
\end{tabular}
\end{table}

\begin{table}[htbp]
\centering
\caption{Datasets used to train the SFT Classifier.}
\vspace{1mm}
\label{tab:sft-datasets}
\small
\begin{tabular}{@{} l l l @{}}
\toprule
Dataset & Task type & Note \\
\midrule
BBH\_tiny \citep{suzgun2022bbh} & exact match & 150-question random sample of BBH \\
MMLU\_Redux\_tiny \citep{gema2024mmlu} & multiple choice & 150-question random sample of MMLU-Redux \\
HumanEval \citep{chen2021humaneval} & coding & 114-question sample of HumanEval \\
\bottomrule
\end{tabular}
\end{table}

\begin{table}[htbp]
\centering
\caption{Datasets used to build the performance-profile signal.}
\label{tab:perf-profile-datasets}
\small
\begin{tabular}{@{} l l l @{}}
\toprule
Dataset & Task type & Size \\
\midrule
ARC-Challenge \citep{clark2018arc} & multiple choice & 1{,}172 \\
BBH \citep{suzgun2022bbh} & exact match & 1{,}000 \\
MMLU-Redux \citep{gema2024mmlu} & multiple choice & 1{,}000 \\
GPQA-Diamond \citep{rein2023gpqa} & multiple choice & 99 \\
MATH \citep{hendrycks2021math} & exact match & 956 \\
MedQA \citep{jin2020medqa} & multiple choice & 637 \\
PopQA \citep{mallen2023popqa} & F1 match & 1{,}000 \\
HumanEval \citep{chen2021humaneval} & coding & 114 \\
CoCoNot \citep{brahman2024coconot} & refusal judgment & 1{,}000 \\
Human-interest \citep{feng2026moco,feng2026one} & reward-model score & 400 \\
\bottomrule
\end{tabular}
\end{table}

\begin{table}[htbp]
\centering
\caption{Datasets used in the out-of-distribution generalization
experiment.}
\label{tab:ood-datasets}
\small
\begin{tabular}{@{} l l l @{}}
\toprule
Dataset & Task type & Size \\
\midrule
GPQA-Diamond \citep{rein2023gpqa} & multiple choice & 99 \\
CoCoNot \citep{brahman2024coconot} & refusal judgment & 1{,}000 \\
\bottomrule
\end{tabular}
\end{table}

\section{Round-by-Round Malicious Model Detection}
\label{app:malicious_grid}

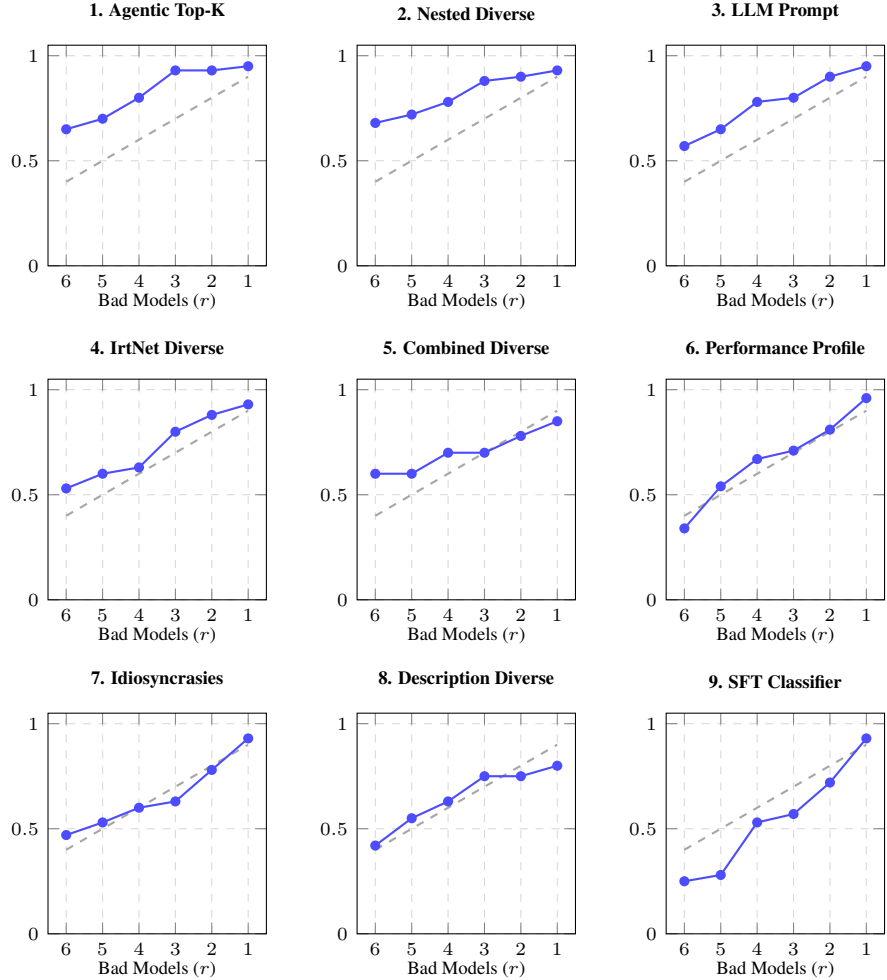
\begin{figure}[htbp]
    \centering
    \begin{tikzpicture}
        \begin{groupplot}[
            group style={
                group size=3 by 3,
                horizontal sep=1.2cm,
                vertical sep=1.5cm
            },
            width=0.32\linewidth,
            height=4.5cm,
            ymin=0, ymax=1.05,
            xmin=0.5, xmax=6.5,
            x dir=reverse, % Matches the UI screenshot (6 bad -> 1 bad)
            xtick={6,5,4,3,2,1},
            ytick={0, 0.5, 1},
            xlabel={Bad Models ($r$)},
            xlabel style={yshift=1ex, font=\scriptsize},
            tick label style={font=\scriptsize},
            grid=major,
            grid style={dashed, gray!30},
            every axis plot/.append style={thick, mark size=1.5pt}
        ]
        
        % Macro for the theoretical random baseline (Dashed gray)
        \def\addrandombaseline{
            \addplot[color=gray!70, dashed, mark=none] coordinates {
                (6, 0.4) (5, 0.5) (4, 0.6) (3, 0.7) (2, 0.8) (1, 0.9)
            };
        }

        % 1. Agentic Top-K
        \nextgroupplot[title={\scriptsize \textbf{1. Agentic Top-K}}]
        \addrandombaseline
        \addplot[color=blue!70!white, mark=*] coordinates {
            (6, 0.65) (5, 0.70) (4, 0.80) (3, 0.93) (2, 0.93) (1, 0.95)
        };

        % 2. Nested Diverse
        \nextgroupplot[title={\scriptsize \textbf{2. Nested Diverse}}]
        \addrandombaseline
        \addplot[color=blue!70!white, mark=*] coordinates {
            (6, 0.68) (5, 0.72) (4, 0.78) (3, 0.88) (2, 0.90) (1, 0.93)
        };

        % 3. LLM Prompt
        \nextgroupplot[title={\scriptsize \textbf{3. LLM Prompt}}]
        \addrandombaseline
        \addplot[color=blue!70!white, mark=*] coordinates {
            (6, 0.57) (5, 0.65) (4, 0.78) (3, 0.80) (2, 0.90) (1, 0.95)
        };

        % 4. IrtNet Diverse
        \nextgroupplot[title={\scriptsize \textbf{4. IrtNet Diverse}}]
        \addrandombaseline
        \addplot[color=blue!70!white, mark=*] coordinates {
            (6, 0.53) (5, 0.60) (4, 0.63) (3, 0.80) (2, 0.88) (1, 0.93)
        };

        % 5. Combined Diverse
        \nextgroupplot[title={\scriptsize \textbf{5. Combined Diverse}}]
        \addrandombaseline
        \addplot[color=blue!70!white, mark=*] coordinates {
            (6, 0.60) (5, 0.60) (4, 0.70) (3, 0.70) (2, 0.78) (1, 0.85)
        };

        % 6. Performance Profile
        \nextgroupplot[title={\scriptsize \textbf{6. Performance Profile}}]
        \addrandombaseline
        \addplot[color=blue!70!white, mark=*] coordinates {
            (6, 0.34) (5, 0.54) (4, 0.67) (3, 0.71) (2, 0.81) (1, 0.96)
        };

        % 7. Idiosyncrasies
        \nextgroupplot[title={\scriptsize \textbf{7. Idiosyncrasies}}]
        \addrandombaseline
        \addplot[color=blue!70!white, mark=*] coordinates {
            (6, 0.47) (5, 0.53) (4, 0.60) (3, 0.63) (2, 0.78) (1, 0.93)
        };

        % 8. Description Diverse
        \nextgroupplot[title={\scriptsize \textbf{8. Description Diverse}}]
        \addrandombaseline
        \addplot[color=blue!70!white, mark=*] coordinates {
            (6, 0.42) (5, 0.55) (4, 0.63) (3, 0.75) (2, 0.75) (1, 0.80)
        };
        
        % 9. SFT Classifier
        \nextgroupplot[title={\scriptsize \textbf{9. SFT Classifier}}]
        \addrandombaseline
        \addplot[color=blue!70!white, mark=*] coordinates {
            (6, 0.25) (5, 0.28) (4, 0.53) (3, 0.57) (2, 0.72) (1, 0.93)
        };

        \end{groupplot}
    \end{tikzpicture}
    \vspace{0.5em}
    \caption{Round-by-round robustness of every selection method against the theoretical \emph{Random} baseline (dashed gray line). The $y$-axis is the fraction of clean models in the final selected team, and the $x$-axis shows the number of malicious models in the 10-candidate pool. Methods that fall below the dashed line are actively selecting malicious models at a higher rate than blind chance.}
    \label{fig:malicious_grid_full}
\end{figure}

Figure~\ref{fig:malicious_grid_full} provides the detailed, round-by-round trajectory for every selection method as the number of malicious models in the 10-candidate pool increases from 1 to 6.

\section{LLM-Prompt Template}
\label{app:prompt}

For the \emph{LLM-prompt} selection method in Family V, we query the instruction-tuned judge to elicit a team-level judgment in a single shot. The exact prompt template provided to the model is shown below. Variables such as the total pool size (\texttt{\{P\}}), the enumerated candidate list (\texttt{\{LIST\_OF\_MODELS\}}), and the target team size (\texttt{\{N\}}) are populated dynamically per experiment.

\vspace{1em}
\begin{quote}
\ttfamily
You are selecting models to collaborate on a task. Below is a pool of \{P\} candidate models, each described by its fine-tuning specialty:

\{LIST\_OF\_MODELS\}

Select exactly \{N\} models from the list above that would work well together as a collaborating team --- consider both individual strengths and complementary diversity between them.

Respond with ONLY a comma-separated list of the chosen indices, and nothing else. Do not explain your reasoning. Do not repeat an index. Example format for selecting 3 models: 2, 5, 7
\end{quote}
\vspace{1em}

\noindent The enumerated \texttt{\{LIST\_OF\_MODELS\}} is formatted as a newline-separated list, where each line consists of the candidate's integer index and its one-line natural language description (e.g., \texttt{0: [Description of model 0]}). To ensure the selector's outputs are perfectly reproducible, the prompt is evaluated using greedy decoding.

\section{Agentic Interview Protocol}
\label{app:agentic-interview}

Here is the full protocol behind the \textsc{Agentic top-$k$} selector (Section~\ref{sec:methodology}, Family V), including how questions are generated and how the initial verdict is produced and corrected.

\subsection{Turn-by-turn question generation}
\label{app:agentic-interview:questions}

The interviewer (\texttt{Qwen2.5-7B-Instruct}) conducts a fixed four-turn interview with each candidate independently. Each turn, it is shown the transcript so far and asked to pick one of five axes not yet covered (step-by-step reasoning, code/programming ability, factual knowledge, calibrated honesty, open-ended conversational quality) and pose one new question probing it, under the following instruction (reproduced verbatim, with dynamic slots in angle brackets):

\vspace{1em}
\begin{quote}\small\ttfamily
You are interviewing a candidate AI model to decide whether to hire it onto a collaborative team. Interview so far: \textnormal{$\langle$transcript so far$\rangle$}. Candidate axes to probe across the whole interview: \textnormal{$\langle$axis list$\rangle$}. Axes already covered so far: \textnormal{$\langle\cdot\rangle$}. Axes NOT yet covered: \textnormal{$\langle\cdot\rangle$}. Pick ONE axis from the NOT yet covered list above and ask exactly ONE new question that probes it. This candidate is likely a competent instruction-tuned model, so a generic ``explain concept X'' or ``describe a time you did Y'' question will get a generically good answer from almost any such model and won't tell us anything useful. Instead, design the question to actively expose a gap or error if one exists: give it a subtly flawed premise to catch, a genuine edge case, a problem with a specific checkable right answer, or a request that tempts an overconfident wrong answer instead of an honest ``I don't know.'' Respond in EXACTLY this format, nothing else: AXIS: \textnormal{$\langle$axis name$\rangle$} QUESTION: \textnormal{$\langle$the question$\rangle$}
\end{quote}
\vspace{1em}

The interviewer generates both the axis choice and the question greedily (temperature $0$, for reproducibility); the candidate then answers as part of a genuine multi-turn conversation (full history carried forward each turn), sampled at temperature $0.7$, top-$p$ $0.9$, up to $512$ new tokens. Axis and question are parsed from an \texttt{AXIS: ... QUESTION: ...} format.

\subsection{Initial verdict}
\label{app:agentic-interview:verdict}

After four turns, the interviewer reads the complete transcript and issues a verdict under an explicit rubric:

\vspace{1em}
\begin{quote}\small\ttfamily
Rate this candidate's overall quality as a hire on a scale of 1-10. Most instruction-tuned models can produce fluent, well-structured answers. Do not reward fluency or length by itself. Grade strictly against these bands, anchored on correctness and depth relative to what a knowledgeable human expert would say: 9-10: Answers are technically flawless, catch subtleties or edge cases a generic competent model would miss, and any claimed uncertainty is honest and warranted. 7-8: Answers are correct and well-organized but stay at a generic, textbook level. 5-6: At least one answer has a real inaccuracy, an unsupported claim stated with false confidence, or misses the point of the question. 3-4: Multiple answers are wrong, evasive, or fail to engage with what was actually asked. 1-2: Answers are incoherent, largely incorrect, or the candidate asserts obviously false claims confidently.
\end{quote}
\vspace{1em}

The verdict is parsed strictly as (\texttt{SCORE: <1--10>} / \texttt{REASON: <...>}).

\subsection{Correcting for score compression}
\label{app:agentic-interview:compression}

Independent, per-candidate verdicts compress toward a lenient, narrow band in practice: on Pool~1, all ten candidates initially scored $9$ or $9.5$ out of $10$ despite visibly different answer quality, even after the rubric-anchoring above. We correct this with a \emph{comparative} re-scoring pass: batches of up to $10$ candidates' full transcripts are shown to the interviewer side-by-side, anonymized as Candidate A/B/C/\ldots, under an instruction that explicitly forbids defaulting to identical scores:

\vspace{1em}
\begin{quote}\small\ttfamily
Compare them against each other directly --- do not score each one in isolation. [\ldots] Critically: you MUST differentiate between candidates whose answers differ in correctness, depth, or honesty about uncertainty. Giving multiple candidates the identical score is only acceptable if their answers are genuinely indistinguishable in quality --- re-read the transcripts and look for a concrete difference (a missed edge case, an unsupported claim, more precise reasoning) before doing so.
\end{quote}
\vspace{1em}

This is the score actually used by \textsc{Agentic top-$k$}; the initial per-candidate verdict above is only a Phase-1 draft.

\subsection{Tie-break refinement}
\label{app:agentic-interview:tiebreak}

If, after averaging, multiple candidates still land on the exact same score, a final pass forces a strict ranking within each tied group (no two candidates may share a rank) and spreads that group's score upward into a fractional span according to rank, capped strictly below both the next distinct score already present in the data and the scale's ceiling of $10$. This ensures no two final scores can collide, and no spread can push a score out of range.

\begin{table}[htbp]
\centering
\caption{Pool 1: 10 Qwen2.5-7B checkpoints, differing only in fine-tuning corpus}
\label{tab:pool1-models}
\small
\begin{tabular}{@{} l l @{}}
\toprule
\textbf{Variant} & \textbf{Specialization} \\
\midrule
\href{https://huggingface.co/bunsenfeng/yuru_qw_code_alpaca}{\nolinkurl{yuru_qw_code_alpaca}} & Code generation and programming tasks \\
\href{https://huggingface.co/bunsenfeng/yuru_qw_cot}{\nolinkurl{yuru_qw_cot}} & Chain-of-thought, step-by-step reasoning \\
\href{https://huggingface.co/bunsenfeng/yuru_qw_flan_v2}{\nolinkurl{yuru_qw_flan_v2}} & Diverse instruction-following across task types \\
\href{https://huggingface.co/bunsenfeng/yuru_qw_gemini_alpaca}{\nolinkurl{yuru_qw_gemini_alpaca}} & General-purpose instruction-following \\
\href{https://huggingface.co/bunsenfeng/yuru_qw_lima}{\nolinkurl{yuru_qw_lima}} & High-quality, concise, well-aligned responses \\
\href{https://huggingface.co/bunsenfeng/yuru_qw_oasst1}{\nolinkurl{yuru_qw_oasst1}} & Open-ended assistant-style conversation \\
\href{https://huggingface.co/bunsenfeng/yuru_qw_open_orca}{\nolinkurl{yuru_qw_open_orca}} & Complex reasoning and detailed explanations \\
\href{https://huggingface.co/bunsenfeng/yuru_qw_science}{\nolinkurl{yuru_qw_science}} & Scientific knowledge and STEM reasoning \\
\href{https://huggingface.co/bunsenfeng/yuru_qw_sharegpt}{\nolinkurl{yuru_qw_sharegpt}} & Conversational, general-purpose \\
\href{https://huggingface.co/bunsenfeng/yuru_qw_wizardlm}{\nolinkurl{yuru_qw_wizardlm}} & Complex, multi-step instruction-following \\
\bottomrule
\end{tabular}
\end{table}

\begin{table}[htbp]
\centering
\caption{Pool 2: 32 models distilled from independently contributed systems. Base architecture is read from the original repository name where stated; ``unspecified'' means neither the name nor the source paper states it, and we did not assume one.} 
\label{tab:pool2-models}
\scriptsize
\begin{tabular}{@{} l >{\raggedright\arraybackslash}p{0.32\linewidth} l >{\raggedright\arraybackslash}p{0.30\linewidth} @{}}
\toprule
\textbf{ID} & \textbf{Original source} & \textbf{Size (B)} & \textbf{Specialization} \\
\midrule
\href{https://huggingface.co/bunsenfeng/parti_0_full}{\nolinkurl{parti_0}}  & \href{https://huggingface.co/chtmp223/Qwen2.5-7B-CLIPPER}{\nolinkurl{chtmp223/Qwen2.5-7B-CLIPPER}} & 7 (Qwen2.5) & Narrative claim verification against full book text \\
\href{https://huggingface.co/bunsenfeng/parti_1_full}{\nolinkurl{parti_1}}  & \href{https://huggingface.co/chengq9/ToolRL-Qwen2.5-3B}{\nolinkurl{chengq9/ToolRL-Qwen2.5-3B}} & 3 (Qwen2.5) & RL-trained for tool use and parameter filling \\
\href{https://huggingface.co/bunsenfeng/parti_2_full}{\nolinkurl{parti_2}}  & \href{https://huggingface.co/AgentFlow/agentflow-planner-7b}{\nolinkurl{AgentFlow/agentflow-planner-7b}} & 7 (unspecified) & Online agent planning via Flow-GRPO \\
\href{https://huggingface.co/bunsenfeng/parti_3_full}{\nolinkurl{parti_3}}  & \href{https://huggingface.co/nanami/ladder-last16L-llama3.1-8b-instruct-sft}{\nolinkurl{nanami/ladder-last16L-llama3.1-8b-instruct-sft}} & 8 (Llama-3.1) & Efficient transformer architecture research \\
\href{https://huggingface.co/bunsenfeng/parti_4_full}{\nolinkurl{parti_4}}  & \href{https://huggingface.co/viswavi/qwen2.5-rlcf}{\nolinkurl{viswavi/qwen2.5-rlcf}} & 7 (Qwen2.5) & Instruction-following via preference tuning on WildChecklists \\
\href{https://huggingface.co/bunsenfeng/parti_5_full}{\nolinkurl{parti_5}}  & \href{https://huggingface.co/milli19/promptmii-llama3.1-8b-instruct}{\nolinkurl{milli19/promptmii-llama3.1-8b-instruct}} & 8 (Llama-3.1) & RL-trained to induce instructions \\
\href{https://huggingface.co/bunsenfeng/parti_6_full}{\nolinkurl{parti_6}}  & \href{https://huggingface.co/Zhengping/conditional-probability-regression}{\nolinkurl{Zhengping/conditional-probability-regression}} & 15 (Qwen2.5) & Conditional-probability estimation in everyday scenarios \\
\href{https://huggingface.co/bunsenfeng/parti_7_full}{\nolinkurl{parti_7}}  & \href{https://huggingface.co/yale-nlp/MDCure-Qwen2-7B-Instruct}{\nolinkurl{yale-nlp/MDCure-Qwen2-7B-Instruct}} & 7 (Qwen2) & Multi-document QA, summarization, coreference resolution \\
\href{https://huggingface.co/bunsenfeng/parti_8_full}{\nolinkurl{parti_8}}  & \href{https://huggingface.co/GritLM/GritLM-7B}{\nolinkurl{GritLM/GritLM-7B}} & 7 (unspecified) & Joint generative and embedding representation model \\
\href{https://huggingface.co/bunsenfeng/parti_9_full}{\nolinkurl{parti_9}}  & \href{https://huggingface.co/lime-nlp/Qwen2.5-7B-Instruct-SUM10}{\nolinkurl{lime-nlp/Qwen2.5-7B-Instruct-SUM10}} & 7 (Qwen2.5) & Uncertainty-aware abstention \\
\href{https://huggingface.co/bunsenfeng/parti_10_full}{\nolinkurl{parti_10}} & \href{https://huggingface.co/geyang627/care-chinese-gemma2-9b}{\nolinkurl{geyang627/care-chinese-gemma2-9b}} & 9 (Gemma-2) & Chinese cultural awareness (CARE dataset) \\
\href{https://huggingface.co/bunsenfeng/parti_11_full}{\nolinkurl{parti_11}} & \href{https://huggingface.co/bespokelabs/Bespoke-Stratos-7B}{\nolinkurl{bespokelabs/Bespoke-Stratos-7B}} & 7 (unspecified) & Hallucination detection \\
\href{https://huggingface.co/bunsenfeng/parti_12_full}{\nolinkurl{parti_12}} & \href{https://huggingface.co/kangdawei/Llama-3.1-8B-Instruct-GenderNeutral-Finetuned}{\nolinkurl{kangdawei/Llama-3.1-8B-Instruct-GenderNeutral-Finetuned}} & 8 (Llama-3.1) & Gender-bias mitigation \\
\href{https://huggingface.co/bunsenfeng/parti_13_full}{\nolinkurl{parti_13}} & \href{https://huggingface.co/DeepRetrieval/DeepRetrieval-PubMed-3B-Llama}{\nolinkurl{DeepRetrieval/DeepRetrieval-PubMed-3B-Llama}} & 3 (Llama) & Query rewriting for retrieval (BM25 / search engines) \\
\href{https://huggingface.co/bunsenfeng/parti_14_full}{\nolinkurl{parti_14}} & \href{https://huggingface.co/yale-nlp/MDCure-Qwen2-1.5B-Instruct}{\nolinkurl{yale-nlp/MDCure-Qwen2-1.5B-Instruct}} & 1.5 (Qwen2) & Multi-document tasks, robust cross-domain generalization \\
\href{https://huggingface.co/bunsenfeng/parti_15_full}{\nolinkurl{parti_15}} & \href{https://huggingface.co/Zhaoxuan/PUGC-Mistral-DPO}{\nolinkurl{Zhaoxuan/PUGC-Mistral-DPO}} & 7 (Mistral) & Preference alignment on user-generated content via DPO \\
\href{https://huggingface.co/bunsenfeng/parti_16_full}{\nolinkurl{parti_16}} & \href{https://huggingface.co/jwhj/Qwen2.5-Math-1.5B-OREO}{\nolinkurl{jwhj/Qwen2.5-Math-1.5B-OREO}} & 1.5 (Qwen2.5) & Math reasoning via offline reasoning optimization \\
\href{https://huggingface.co/bunsenfeng/parti_17_full}{\nolinkurl{parti_17}} & \href{https://huggingface.co/PeterJinGo/SearchR1-nq_hotpotqa_train-qwen2.5-7b-em-ppo}{\nolinkurl{PeterJinGo/SearchR1-nq_hotpotqa_train-qwen2.5-7b-em-ppo}} & 7 (Qwen2.5) & Agentic search via PPO reinforcement learning \\
\href{https://huggingface.co/bunsenfeng/parti_18_full}{\nolinkurl{parti_18}} & \href{https://huggingface.co/gasolsun/DynamicRAG-8B}{\nolinkurl{gasolsun/DynamicRAG-8B}} & 8 (unspecified) & Dynamic reranking agent for retrieved documents \\
\href{https://huggingface.co/bunsenfeng/parti_19_full}{\nolinkurl{parti_19}} & \href{https://huggingface.co/LLM360/guru-7B}{\nolinkurl{LLM360/guru-7B}} & 7 (unspecified) & General reasoning (GURU dataset) \\
\href{https://huggingface.co/bunsenfeng/parti_20_full}{\nolinkurl{parti_20}} & \href{https://huggingface.co/spiral-rl/Spiral-Qwen3-4B}{\nolinkurl{spiral-rl/Spiral-Qwen3-4B}} & 4 (Qwen3) & Multi-agent self-play (Poker, Tic-Tac-Toe) transferring to math/logic \\
\href{https://huggingface.co/bunsenfeng/parti_21_full}{\nolinkurl{parti_21}} & \href{https://huggingface.co/uclanlp/brief-pro}{\nolinkurl{uclanlp/brief-pro}} & 4 (Llama-3.2) & Lightweight evidence compressor for in-context RAG \\
\href{https://huggingface.co/bunsenfeng/parti_22_full}{\nolinkurl{parti_22}} & \href{https://huggingface.co/Rakancorle1/PolicyGuard4B}{\nolinkurl{Rakancorle1/PolicyGuard4B}} & 4 (unspecified) & Guardrail detecting policy violations in web-agent trajectories \\
\href{https://huggingface.co/bunsenfeng/parti_23_full}{\nolinkurl{parti_23}} & \href{https://huggingface.co/ReasoningTransferability/UniReason-Qwen3-14B-RL}{\nolinkurl{ReasoningTransferability/UniReason-Qwen3-14B-RL}} & 14 (Qwen3) & RL-GRPO math reasoning; transferability to general language tasks \\
\href{https://huggingface.co/bunsenfeng/parti_24_full}{\nolinkurl{parti_24}} & \href{https://huggingface.co/ypwang61/One-Shot-RLVR-Qwen2.5-Math-1.5B-pi1}{\nolinkurl{ypwang61/One-Shot-RLVR-Qwen2.5-Math-1.5B-pi1}} & 1.5 (Qwen2.5) & One-example RLVR training, overfitting robustness \\
\href{https://huggingface.co/bunsenfeng/parti_25_full}{\nolinkurl{parti_25}} & \href{https://huggingface.co/sunblaze-ucb/Qwen2.5-3B-Intuitor-MATH-1EPOCH}{\nolinkurl{sunblaze-ucb/Qwen2.5-3B-Intuitor-MATH-1EPOCH}} & 3 (Qwen2.5) & Self-certainty reward (RLIF) on MATH \\
\href{https://huggingface.co/bunsenfeng/parti_26_full}{\nolinkurl{parti_26}} & \href{https://huggingface.co/yale-nlp/Qwen3-8B-SciLit-01}{\nolinkurl{yale-nlp/Qwen3-8B-SciLit-01}} & 8 (Qwen3) & Scientific and STEM reasoning \\
\href{https://huggingface.co/bunsenfeng/parti_27_full}{\nolinkurl{parti_27}} & \href{https://huggingface.co/OpenThoughts/OpenThinker3-7B}{\nolinkurl{OpenThoughts/OpenThinker3-7B}} & 7 (unspecified) & Open-source reasoning (OpenThoughts dataset) \\
\href{https://huggingface.co/bunsenfeng/parti_28_full}{\nolinkurl{parti_28}} & \href{https://huggingface.co/l3lab/L1-Qwen-1.5B-Exact}{\nolinkurl{l3lab/L1-Qwen-1.5B-Exact}} & 1.5 (Qwen) & Length-controlled reasoning via RL \\
\href{https://huggingface.co/bunsenfeng/parti_29_full}{\nolinkurl{parti_29}} & \href{https://huggingface.co/fangwu97/DeepSearch-1.5B}{\nolinkurl{fangwu97/DeepSearch-1.5B}} & 1.5 (unspecified) & Reasoning enhanced by Monte Carlo Tree Search \\
\href{https://huggingface.co/bunsenfeng/parti_30_full}{\nolinkurl{parti_30}} & \href{https://huggingface.co/allegrolab/hubble-8b-500b-toks-standard-hf}{\nolinkurl{allegrolab/hubble-8b-500b-toks-standard-hf}} & 8 (Llama, from scratch) & Memorization study, 3.7$\times$ Chinchilla-optimal training \\
\href{https://huggingface.co/bunsenfeng/parti_31_full}{\nolinkurl{parti_31}} & \href{https://huggingface.co/fcyin/llama2_7B_base_lofit_truthfulqa}{\nolinkurl{fcyin/llama2_7B_base_lofit_truthfulqa}} & 7 (Llama-2) & TruthfulQA tuning via attention-head editing (LoFiT) \\
\bottomrule
\end{tabular}
\end{table}

\end{document}

%% file: math_commands.tex
\usepackage{amsmath,amsfonts,bm}

\def\eqref#1{equation~\ref{#1}}
\def\1{\bm{1}}

\DeclareMathAlphabet{\mathsfit}{\encodingdefault}{\sfdefault}{m}{sl}
\SetMathAlphabet{\mathsfit}{bold}{\encodingdefault}{\sfdefault}{bx}{n}